\documentclass[times]{TRR}

\usepackage{moreverb,url}
\usepackage{xurl}
\usepackage[colorlinks,bookmarksopen,bookmarksnumbered,citecolor=red,urlcolor=red]{hyperref}

\usepackage{subcaption}
\usepackage{tabularx}

\newcommand\BibTeX{{\rmfamily B\kern-.05em \textsc{i\kern-.025em b}\kern-.08em
T\kern-.1667em\lower.7ex\hbox{E}\kern-.125emX}}

\def\volumeyear{2025}

\begin{document}

\runninghead{Rahmani et al.}

\title{Automated Vehicles at Unsignalized Intersections: Safety and Efficiency Implications of Mixed Human and Automated Traffic}

\author{Saeed Rahmani\affilnum{1}, Zhenlin (Gavin) Xu\affilnum{1}, Simeon C. Calvert\affilnum{1}, Bart van Arem\affilnum{1}}

\affiliation{\affilnum{1}Department of Transport \& Planning, Faculty of Civil Engineering and Geosciences, Delft University of Technology, 2628 CD Delft, NL}

\corrauth{Saeed Rahmani, s.rahmani@tudelft.nl}

\begin{abstract}
The integration of automated vehicles (AVs) into transportation systems presents an unprecedented opportunity to enhance road safety and efficiency. However, understanding the interactions between AVs and human-driven vehicles (HVs) at intersections remains an open research question. This study aims to bridge this gap by examining behavioral differences and adaptations of AVs and HVs at unsignalized intersections by utilizing two large-scale AV datasets from Waymo and Lyft. By using a systematic methodology, the research identifies and analyzes merging and crossing conflicts by calculating key safety and efficiency metrics, including time to collision (TTC), post-encroachment time (PET), maximum required deceleration (MRD), time advantage (TA), and speed and acceleration profiles.  Through this approach, the study assesses the safety and efficiency implications of these behavioral differences and adaptations for mixed-autonomy traffic. The findings reveal a paradox: while AVs maintain larger safety margins, their conservative behavior can lead to unexpected situations for human drivers, potentially causing unsafe conditions. From a performance point of view, human drivers tend to exhibit more consistent behavior when interacting with AVs versus other HVs, suggesting AVs may contribute to harmonizing traffic flow patterns. Moreover, notable differences were observed between Waymo and Lyft vehicles, which highlights the importance of considering manufacturer-specific AV behaviors in traffic modeling and management strategies for the safe integration of AVs. The processed dataset, as well as the developed algorithms and scripts, are openly published to foster research on AV-HV interactions.
\end{abstract}

\maketitle

\section{Introduction}
Automated vehicles (AVs) are widely recognized as a transformative innovation in transportation, with the potential to improve traffic safety and efficiency \cite{tafidis2022safety, cohen2019automated, aittoniemi2022evidence, milakis2017policy}. In recent years, automated vehicles have rapidly evolved from concept to reality, with companies like Waymo and Motional already operating robotaxi fleets across U.S. cities, and Tesla and Zoox preparing to launch similar services. \cite{reuters_tesla_robotaxi_2024, bloomberg_zoox_robotaxi_2024}. As AVs transition from prototypes to real-world products, understanding their interactions with human drivers and impacts on transportation systems is essential for their safe integration, effective regulation, and building public trust in these technologies \cite{rahmati2019influence, jiao2024beyond, albano2024drivers}.

\textcolor{black}{The growing availability of real-world AV datasets in recent years has opened up unprecedented opportunities to study the dynamics and interactions between automated and human-driven vehicles. Existing research has examined both the operational characteristics of AVs navigating alongside conventional vehicles \cite{hu2023autonomous, wen2022characterizing} and the behavioral adaptations of human drivers when interacting with AVs \cite{rahmati2019influence, ma2024driver, wen2022characterizing, zhang2023impact, wen2023modeling, li2023large}. While these studies provide valuable insights, they have mainly been limited to one-dimensional, longitudinal interactions, also known as car-following scenarios. However, longitudinal interactions alone cannot fully capture the complexity of real-world AV-HV dynamics, such as those at unsignalized intersections. Such scenarios involve complex multi-directional interactions that require advanced decision-making, negotiation, and prediction of other agents’ intentions \cite{zhao2020two, rahmani2023bi}. As a result, the dynamics in these settings are considerably more challenging and intricate than in car-following situations \cite{boggs2020exploratory, albano2024drivers}. Moreover, existing studies typically rely on a single automated vehicle dataset, limiting the generalizability of their findings and overlooking differences in design philosophies, sensor configurations, and decision-making algorithms among manufacturers.
Therefore, comparative analyses across multiple datasets are essential to uncover how differences in manufacturer-specific AV behaviors affect traffic flow efficiency and safety. Such insights are necessary for developing realistic models and regulatory policies that support the safe and efficient deployment of automated vehicles.}

\textcolor{black}{In response to these gaps, the present study conducts a comprehensive analysis of AV-HV interactions at unsignalized intersections using two large-scale AV datasets. By examining both merging and crossing conflicts, and utilizing various safety and efficiency metrics, we aim to capture the intricate dynamics of these interactions and shed light on the potential safety and efficiency implications of mixed-human-automated traffic.} The main contributions of this study are summarized as follows:
\begin{itemize}
    \item \textit{Unravelling AV-HV Interactions at Unsignalized Intersections}: We provide a thorough examination of the interactions between AVs and HVs at unsignalized intersections. To our knowledge, this is the first study focusing on behavioral differences between AVs and HVs at unsignalized intersections.
    \item \textit{Utilizing Multiple Real-World Datasets}: Unlike previous studies that have primarily relied on a single dataset, we utilize two large-scale AV datasets, which allows us to examine the potential behavioral variations between different AV platforms and the possible adaptations of human drivers to AVs with different appearances and driving styles.
    \item \textit{Novel Metrics for Analyzing Driving Behavior:} This study introduces Maximum Required Deceleration (MRD) and the distribution of Time Advantage (TA) as novel metrics to analyze driving styles and aggressiveness. MRD captures peak braking demand during conflicts, while TA assesses positional advantages and negotiation dynamics. These metrics could offer new insights into interactions at unsignalized intersections. 
    \item \textit{Provision of a High-Quality Conflict Dataset for Unsignalized Intersections}: We provide a meticulously processed dataset for unsignalized intersections, including merging and crossing conflicts. By denoising raw data and identifying key interactions, we create a valuable resource to support further research on AV-HV interactions at unsignalized intersections. 
\end{itemize}
Through these contributions, this study provides a better understanding of AV-HV dynamics in complex traffic scenarios and aims to advance the growing body of knowledge on AV-HV interactions. The remainder of the paper is as follows. In the next section, the related studies are reviewed. Next, the methodology is described, including the data preprocessing, scenario selection, and metrics calculations. Finally, the results are presented and discussed.

\section{Related Works}
Previous research on AV-HV interactions generally falls into two main categories: studies exploring the behavioral differences between AVs and HVs, and studies investigating the behavioral adaptation of HVs when interacting with AVs. In the context of mixed-autonomy traffic, ``behavioral adaptation'' refers to the behavioral adjustments made by human drivers when interacting with automated vehicles compared to when they are interacting with other human drivers \cite{soni2022behavioral}. These adaptations may include changes in yielding behavior, gap acceptance, and acceleration or deceleration profiles. In this section, we review these studies and highlight the gaps in the existing literature.

The initial studies utilized ``field data'' to research the behavior of human drivers in mixed-autonomy traffic. \cite{rahmati2019influence} conducted car-following experiments using Texas A\&M University’s automated Chevy Bolt and observed that human drivers felt more comfortable following the AV and kept shorter headways. \citep{soni2022behavioral} performed a Wizard of Oz experiment in the Netherlands focusing on car-following and overtaking behaviors. They found that human drivers can potentially exploit AVs in their interactions by making abrupt merges in front of AVs. \citep{zhao2020field} doubted these findings and argued that drivers’ reactions to autonomous vehicles depend on their subjective trust in AV technologies rather than the actual driving behavior. They categorized humans as AV-believers and AV-skeptics, where only AV-believers followed AVs with lower headways. \citep{ma2024driver} observed a similar pattern in the interactions between HVs and AVs, where only aggressive and moderate drivers showed increased aggressive behavior. Additionally, aggressive drivers were more prone to exploit AVs while driving.

With the availability of ``real-world'' datasets of AVs, such as Waymo Open Dataset \cite{hu2022processing, mei2022waymo}, Lyft Level 5 \cite{houston2021one}, and Argoverse \cite{wilson2023argoverse}, researchers began to characterize the behavior AVs and HVs in mixed-autonomy traffic. \citep{wen2022characterizing} and \citep{zhang2023impact} noted that human drivers following AVs exhibit lower driving volatility, shorter time headways, and higher time-to-collision values. This finding is confirmed by \citep{wen2023modeling} and \citep{li2023large}, who observed that human drivers maintain shorter spacing and time gaps when following AVs compared to human-driven vehicles. On the other side, and focusing on the behavior of AVs among HVs, \citep{hu2023autonomous} found that AVs demonstrate significantly larger time headways than HVs, suggesting higher safety margins. They also noted longer response times for AVs to various stimuli. Despite these relatively consistent findings from real-world datasets, \citep{jiao2024beyond} cast doubt upon these findings and propose alternative explanations for these observed behavioral differences. They highlight potential observation biases in AV-collected data, driver heterogeneity, and distinct driving patterns between AVs and HVs and argue that these behavioral insights from non-behavioral data require further scrutiny. All in all, these studies collectively underscore the behavior adaptation of HVs when interacting with AVs, and the different behavior of AVs compared to human drivers. However, they have mainly focused on car-following scenarios and neglected the intricate and multi-dimensional interactions at urban intersections. 

Recently, a few studies have gone beyond analyzing car-following scenarios. \citep{wen2022characterizing} studied the behavior of AVs and HVs when approaching signalized intersections. However, they only focused on longitudinal behaviors of AVs and HVs before the intersection, such as approaching a queue or starting a maneuver after a traffic light turns green, and did not investigate the interactions of vehicles inside the intersections. \citep{li2023comparative} identified crossing conflicts within the Argoverse 2 dataset \cite{wilson2023argoverse} to study the conflict resolution behavior of AVs and HVs in AV-free and AV-included scenarios. Although this study includes two-dimensional interactions, it does not differentiate between unsignalized intersections and other conflicts, such as vehicles exiting parking areas or situations where one vehicle has a clear right of way. Mixing these scenarios with unsignalized intersections can lead to biased estimations, as they involve different levels of complexity and decision-making. Additionally, the study is limited by its reliance on a single dataset, which questions its generalizability and restricts the ability to investigate behavioral differences across AVs from different companies, and potential different behavioral adaptations of human drivers when facing AVs with different driving styles. Last but not least, the study primarily focuses on crossing maneuvers, without comparing the outcomes of merging and crossing conflicts, which limits a more comprehensive understanding of these different types of interactions.

This study addresses these limitations by utilizing two real-world AV datasets from Waymo \cite{sun2020scalability} and Lyft \cite{houston2021one}, identifying unsignalized intersections with equal right-of-way on all approaches, and analyzing both crossing and merging conflicts. Through these contributions, we offer a more robust and diverse analysis of AV-HV interactions. Utilizing two different datasets allows us to explore not only the behavioral differences between AVs and HVs but also between AVs from different manufacturers. Furthermore, the published processed dataset in this study is the first dataset focusing on unsignalized intersections. Through these contributions, we aim to facilitate targeted research in this domain, such as calibrating microsimulation models \cite{rahmani2023bi, zhao2020two} and the development of safer, more efficient, and socially-compliant autonomous driving systems.

\section{Methodology}
\begin{figure*}[t]
    \centering
    \includegraphics[width=\linewidth]{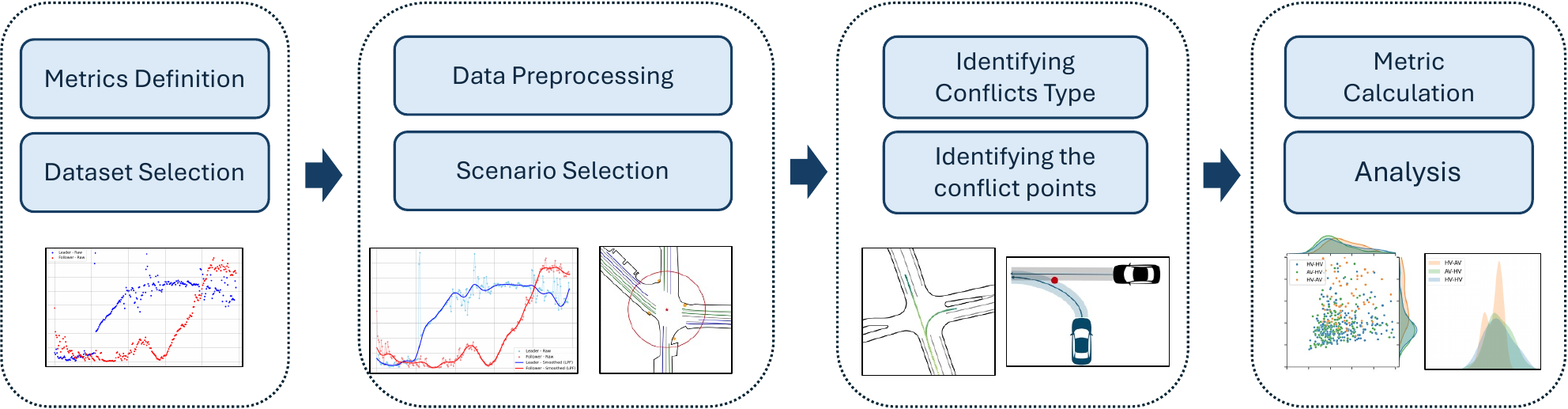}
    \caption{The methodological framework employed in this study, illustrating the systematic process from data selection and preprocessing to conflict identification and behavioral analysis}
    \label{fig:framework}
\end{figure*}

\textcolor{black}{To ensure a robust and reliable analysis of the interactions between automated vehicles and human-driven vehicles, we propose a structured framework that guides the entire process from dataset selection to conflict analysis. The proposed framework, illustrated in Figure~\ref{fig:framework}, begins with defining the analysis metrics and selecting appropriate datasets aligned with the objectives of the study. Next, the selected datasets undergo preparation, which includes developing algorithms for identifying unsignalized intersections from raw datasets and preprocessing the data to mitigate noise and remove outliers. Subsequently, we develop algorithms for detecting and classifying conflicts into crossing or merging interactions, identifying potential conflict points, and calculating metrics that characterize the behaviors and interactions of AVs and HVs. The framework concludes with a comprehensive analysis using the established metrics and statistical methods to ensure robust and reliable findings. Our analysis encompasses both qualitative and statistical comparative analysis. We deliberately adopt the term ``framework'' to highlight the importance of key steps, including data preparation, scenario selection, and conflict identification, which could have substantial impacts on the final results \cite{jiao2024beyond}. This structured approach ensures clarity and consistency and supports the validity and reproducibility of the results. To further promote transparency and reuse, all scripts, algorithms, and processed datasets are made openly available on \url{https://github.com/SaeedRahmani/Unsignalized_AV_HV}.}

\subsection{Definition of Metrics}
\textcolor{black}{Before detailing the data-driven procedures, we first introduce the key metrics used to characterize vehicle interactions and conflict dynamics throughout the study.}

\subsubsection{Time to Collision (TTC):} TTC measures the time remaining before a collision would occur if both vehicles maintained their current speed and trajectory. We use the approach proposed by \citep{albano2024drivers} for calculating two-dimensional TTC for merging and crossing interactions. For crossing scenarios, we calculate TTC as:
\begin{equation}
    TTC_{cross} = \frac{d_f}{v_f}
\end{equation}
For merging scenarios:
\begin{equation}
    TTC_{merge} = \frac{d_f}{v_f - v_l}
\end{equation}
Where $d_f$ is the distance of the follower to the conflict point, and $v_f$ and $v_l$ are the speeds of the following and leading vehicles, respectively. \textcolor{black}{For merging scenarios, we adopted $TTC_{merge}$ as defined by European Regulation 1426/2022 \cite{ciuffo2024interpretation, albano2024drivers}. In this definition, as the vehicles are (almost) parallel at and after the merging point, the relative speed between the leading and following vehicles is used.}
Also, minimum TTC (minTTC), an indicator of conflict severity and risk, is calculated as the minimum value of TTC recorded for an interaction over the duration of the interaction until the second vehicle passes the conflict point.

\subsubsection{Post-Encroachment Time (PET):} PET measures the time difference between the moment the first vehicle leaves the conflict point and the second vehicle arrives at it:
\begin{equation}
    PET = t_f - t_l
\end{equation}
Where $t_f$ is the time the following vehicle arrives at the conflict point, and $t_l$ is the time the leading vehicle leaves the conflict point.

\subsubsection{Maximum Required Deceleration to Avoid Collision (MRD):}
\textcolor{black}{While existing metrics such as TTC and PET capture the timing and spatial proximity of interactions, they do not reflect the intensity of response required from the following vehicle to avoid a collision. This intensity is important because it directly influences the follower’s comfort and the perceived safety of interactions. It also reveals behavioral traits of the leading vehicle, such as aggressiveness or lack of cooperation, which traditional metrics like TTC and PET fail to capture. Accordingly, in this study, we introduce a new metric called maximum required deceleration (MRD). MRD quantifies the peak deceleration that a following vehicle must apply to avoid a potential collision, from the moment a conflict is detected until the leading vehicle clears the conflict point. Higher MRD values indicate increased braking demands on the follower, potentially stemming from aggressive or non-cooperative actions by the leading vehicle. MRD can also serve as a safety metric by representing the criticality of the situation based on the required deceleration needed to avoid potential collisions. This dual role makes MRD a valuable addition to existing metrics, offering insights into both conflict severity and interaction dynamics. MRD is mathematically expressed as:}
\begin{equation}
    MRD = \max \left| \frac{v_{i_f}^2}{2d_{i_f}} \right|
\end{equation}
where \(v_{i_f}\) is the speed of the following vehicle and \(d_{i_f}\) is the distance to the conflict point at each time instance \(i\). \textcolor{black}{Selecting the maximum value is preferable to aggregating the values over time because it highlights the peak braking demand on the follower and indicates the criticality of the situation.} In contrast, an aggregated value might smooth out critical moments, potentially underestimating the severity of the situation.

\subsubsection{Time Advantage (TA):} TA indicates the time differences between the estimated arrival of the first and the second vehicle at the intersection. Time advantage defines which vehicle arrives first at the conflict point assuming that the two vehicles keep their current speed:
\begin{equation}
    TA = \frac{d_f}{v_f} - \frac{d_l}{v_l}
\end{equation}
where \(d_{f}\) represents the distance of the following vehicle to the conflict point, \(d_{l}\) represents the distance of the leading vehicle to the conflict point, \(v_{f}\) is the speed of the following vehicle, and \(v_{l}\) is the speed of the leading vehicle. In this study, we use TA \textit{distributions} to examine how automated and human-driven vehicles negotiate positional advantages during interactions at unsignalized intersections. Specifically, TA distributions are used to identify patterns of assertiveness or yielding behaviors. For instance, TA distributions allow us to detect cases where human drivers exploit automated vehicles by accelerating to gain positional advantage. This novel application of TA enables a detailed analysis of the driving styles and negotiation strategies employed by AVs and human drivers, providing unique insights into mixed-traffic dynamics.

\subsection{Dataset selection and Introduction}
Our criteria for dataset selection included several key characteristics: the availability of long trajectories to capture the behavior of AVs and HVs before and within the intersection, the presence of automated vehicles operating in fully autonomous mode (and not with a human driver), and the availability of comprehensive trajectory data for detailed analysis. Table~\ref{tab:dataset_comparison} presents a comparison of four popular AV datasets. Among these datasets, the nuPlan data is collected by an equipped vehicle driven by a human driver, and the length of scenarios within the Argoverse 2 dataset is 11 seconds. 
Therefore, this study utilizes Lyft Level 5 and Waymo Open Datasets, which provide longer scenarios, allowing us to study both the negotiation phase and the interaction phase among vehicles. Both datasets provide rich information on autonomous vehicle operations in diverse traffic scenarios, making them suitable for our analysis of AV-HV interactions at unsignalized intersections.

\begin{table*}[t]
\caption{Comparison of nuPlan, Waymo, Argoverse 2, and Lyft Level 5 datasets}
\fontsize{11pt}{13pt}\selectfont
\centering
\begin{tabular}{lcccc}
\toprule
\textbf{Dataset} & \textbf{Duration} & \textbf{Trajectory Data} & \textbf{Operation Mode} & \textbf{Length of Scenarios} \\
\midrule
Waymo & 570 hours & Yes (10Hz) & Autonomous & 10 \& 20 seconds \\
Lyft Level 5 & 1000 hours & Yes (10Hz) & Autonomous & 20 seconds \\
nuPlan & 1100 hours & Yes (10Hz) & Human Driver & 25 seconds \\
Argoverse 2 & 763 hours & Yes (10Hz) & Autonomous & 11 seconds \\
\bottomrule
\end{tabular}
\label{tab:dataset_comparison}
\end{table*}

\begin{figure}[b]
    \centering
    \begin{subfigure}[b]{0.45\textwidth}
        \centering
        \includegraphics[width=\textwidth]{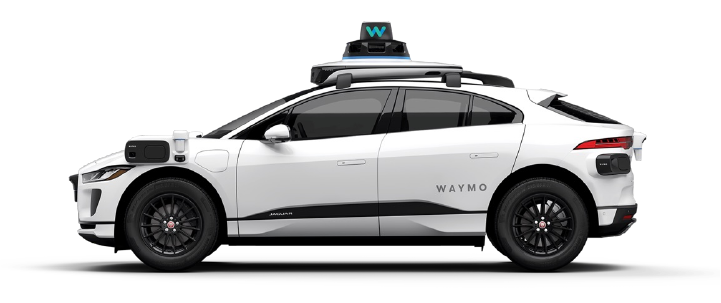}
        \caption{Waymo vehicle - Image source: \cite{roadtoautonomy2024}}
        \label{fig:waymo_veh}
    \end{subfigure}
    \begin{subfigure}[b]{0.38\textwidth}
        \centering
        \includegraphics[width=\textwidth]{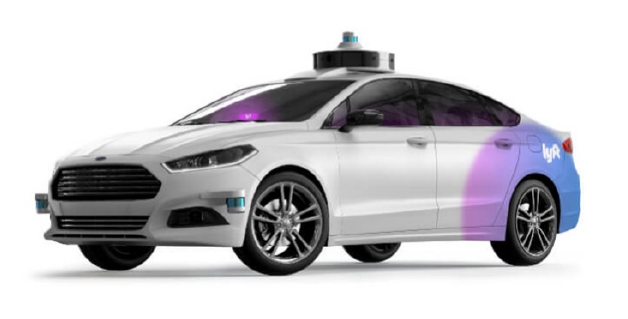}
        \caption{Lyft vehicle - Image source: \cite{gdude2021}}
        \label{fig:lyft_veh}
    \end{subfigure}
    \caption{Visualization of Waymo and Lyft Level 5 vehicles}
    \label{fig:mainfig}
\end{figure}

The Waymo Open Dataset \cite{waymo_open_dataset} provides approximately 570 hours of driving data, collected across six cities, including San Francisco, Phoenix, Mountain View, Los Angeles, Detroit, and Seattle. We utilize version 1.2.1 of the motion dataset (April 2024), which includes 20-second scenarios and provides the labels for the AV, as well as for other agents in each scene. This dataset employs a fleet of Jaguar I-PACE electric SUVs, which are distinguishable from regular vehicles due to their advanced sensor arrays, including roof and side-mounted LiDAR units and multiple high-definition cameras (Figure~\ref{fig:waymo_veh}). This dataset provides high-resolution trajectory data at 10 Hz, and a length of 20 seconds for each scenario, enabling precise analysis of vehicle interactions. 

The Lyft Level 5 dataset \cite{l5kit_dataset} comprises over 1,000 hours of driving data collected in Palo Alto, California. It features a fleet of 20 autonomous vehicles equipped with a sensor suite including LiDAR, radar, and cameras. The vehicles, Ford Fusion models retrofitted with Lyft's autonomous driving technology, are visually distinct from standard human-driven vehicles due to their roof and bumper-mounted sensor arrays (Figure~\ref{fig:lyft_veh}). This dataset also provides detailed trajectory data for both autonomous vehicles and surrounding traffic, including human-driven vehicles, with a temporal resolution of 10 Hz and a length of 20 seconds for each scenario.

\subsection{Dataset Preparation}
The dataset preparation is comprised of two steps: preprocessing the data by removing outliers and reducing noises in the datasets, and identifying relevant scenarios where vehicles interact with each other at an unsignalized intersection. These steps are detailed as follows. 

\subsubsection{Data Smoothing}
\begin{figure}[ht]
    \centering
    \begin{subfigure}[b]{0.47\textwidth}
        \centering
        \includegraphics[width=\textwidth]{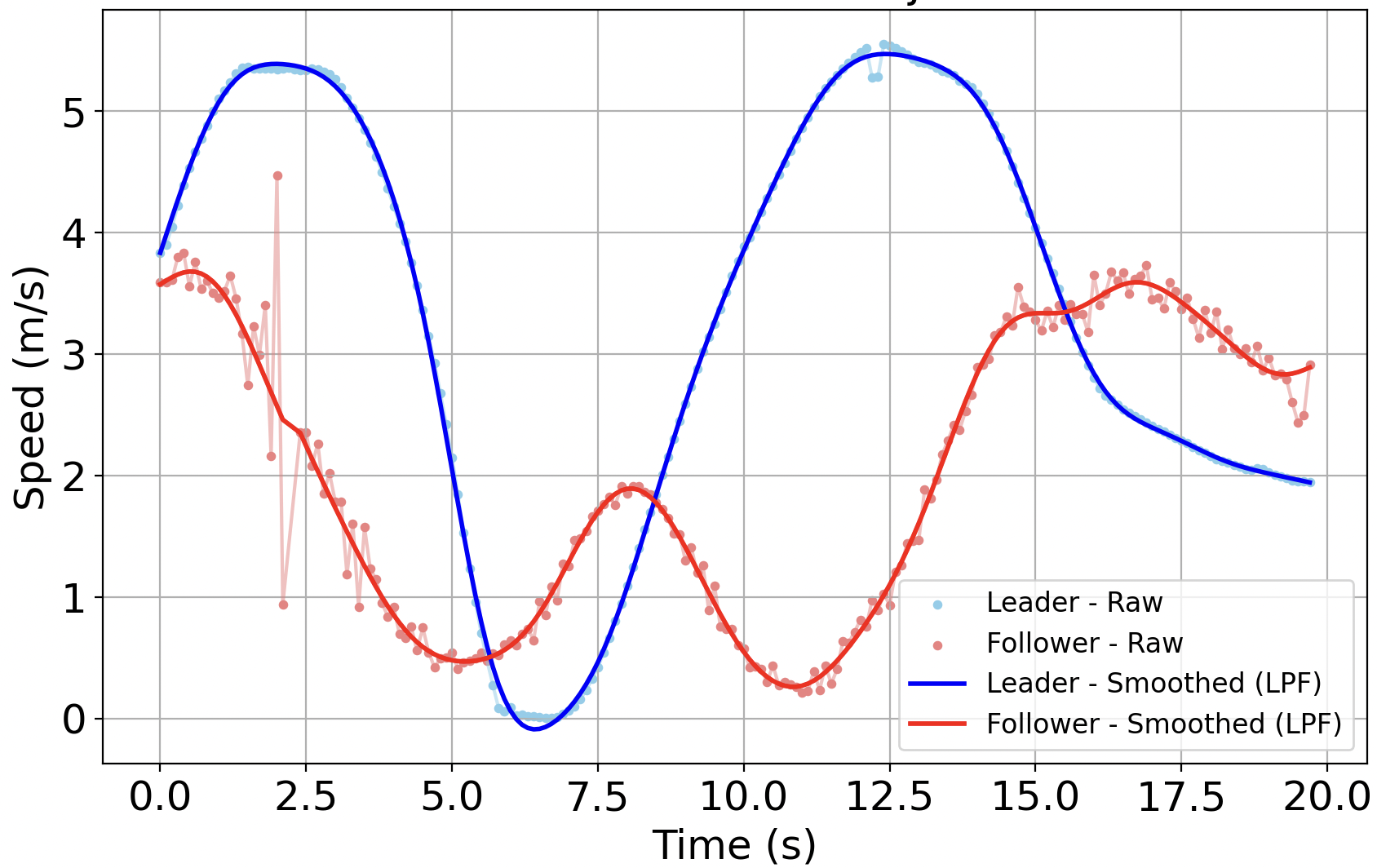}
        \caption{Waymo Dataset}
        \label{fig:waymo_smoothing}
    \end{subfigure}
    \hfill
    \begin{subfigure}[b]{0.47\textwidth}
        \centering
        \includegraphics[width=\textwidth]{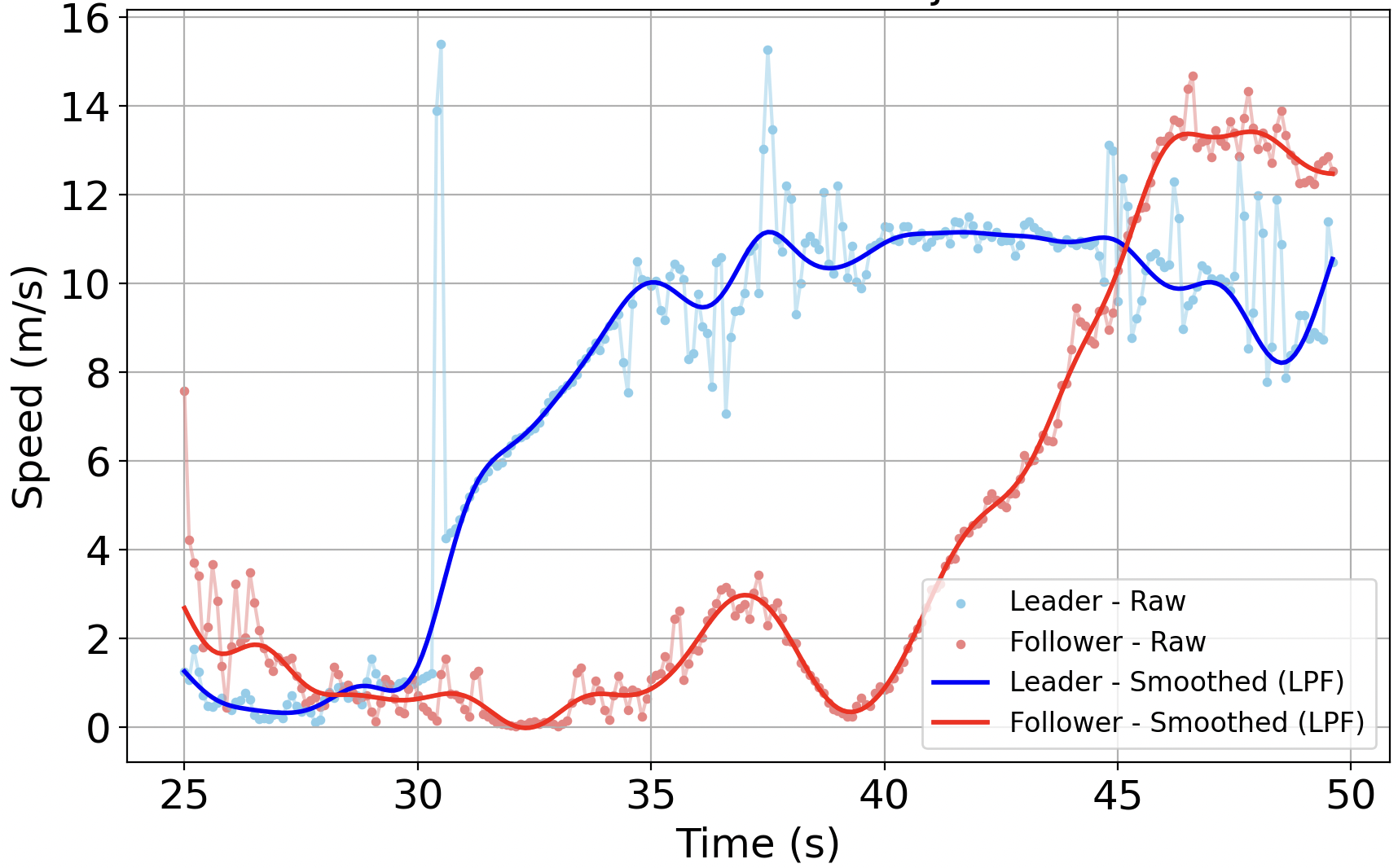}
        \caption{Lyft Dataset}
        \label{fig:lyft_smoothing}
    \end{subfigure}
    \caption{Applying the low pass filter to speed profiles from Waymo and Lyft datasets}
    \label{fig:smoothing}
\end{figure}

We observed different types of noises and outliers in the dataset. Most of the noises were apparently due to measurement errors also reported by \citep{jiao2024beyond}. However, in some cases, we observed high jumps in the speed profiles, which seemed to be due to inaccurate interpolation between scenarios or frames, which was also observed by \citep{li2023large}.  To alleviate such outliers, we first filtered out the data points that required accelerations higher than \(10 \, \text{m/s}^2\) or decelerations lower than \(-10 \, \text{m/s}^2\) and replaced them with the average observed values of the previous five and next five frames. Next, we applied a low-pass Butterworth filter to smooth the high-frequency fluctuations in our velocity data. The low-pass filter is designed to allow low-frequency components of the signal to pass through while attenuating higher-frequency components, effectively smoothing the data \cite{dong2021integrated}. This approach is particularly useful for reducing noise and revealing the underlying trends in our time-series data. It is characterized by its maximally flat frequency response in the passband, which means it doesn't distort the desired part of the signal. The Butterworth filter is characterized by its transfer function:
\begin{equation}
    H(s) = \frac{1}{\sqrt{1 + \left(\frac{s}{\omega_c}\right)^{2n}}}
\end{equation}
where \( s \) is the complex frequency variable, \( \omega_c \) is the cutoff angular frequency, and \( n \) is the filter order. To apply this filter to our discrete-time velocity data, we used a digital implementation by utilizing the SciPy library in Python. The digital Butterworth filter was designed using the following parameters: a cutoff frequency (\( \omega_c \)) of 0.5 Hz, a sampling frequency (\( fs \)) of 10.0 Hz, and a filter order (\( n \)) of 4. The cutoff angular frequency (0.5 Hz) and filter order (4) were determined through domain knowledge, iterative tuning, and empirical testing. We began by exploring a range of cutoff frequencies (0.2–2 Hz) and filter orders (1–6) to balance noise reduction and the preservation of meaningful variations in velocity profiles. After testing various combinations, we selected a cutoff frequency of 0.5 Hz and a filter order of 4, as these values effectively filtered out high-frequency noise while retaining significant speed changes corresponding to events with a period of about 2 seconds or longer, reflecting typical driver reaction times. These choices align with prior studies \cite{dong2021integrated, montanino2013making} and ensured that the analysis focused on capturing critical vehicle dynamics while maintaining data integrity. Figure~\ref{fig:smoothing} shows the results of applying the designed filter and outlier detection to both Waymo and Lyft datasets. As it is presented, the smoothing process has successfully removed the outliers and noises in the datasets, while presenting the main trends.

\subsubsection{Scenario Selection}
\begin{figure*}[t]
    \centering
    \begin{subfigure}[c]{0.49\textwidth}
        \centering
        \includegraphics[width=0.7\textwidth]{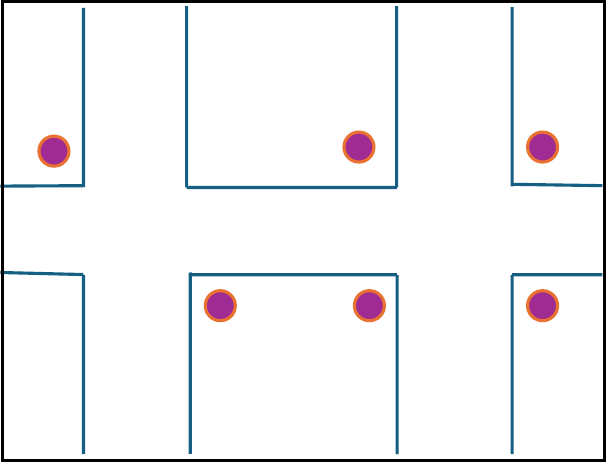}
        \caption{Identifying stop signs}
        \label{fig:all_stops}
    \end{subfigure}
    \hfill
    \begin{subfigure}[c]{0.49\textwidth}
        \centering
        \includegraphics[width=0.7\textwidth]{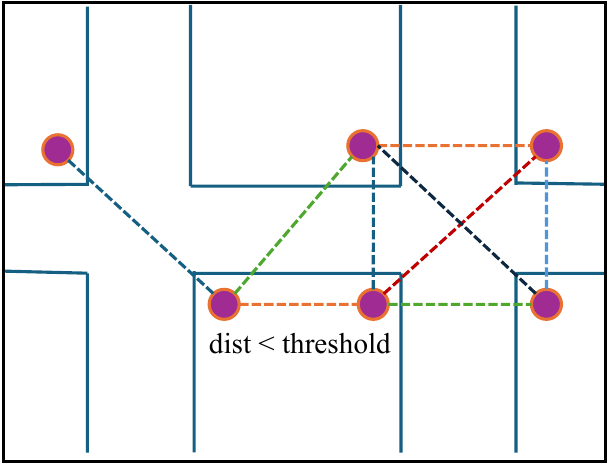}
        \caption{Identifying pairs}
        \label{fig:Pairs}
    \end{subfigure}
    
    \hfill
    \begin{subfigure}[c]{0.49\textwidth}
        \centering
        \includegraphics[width=0.7\textwidth]{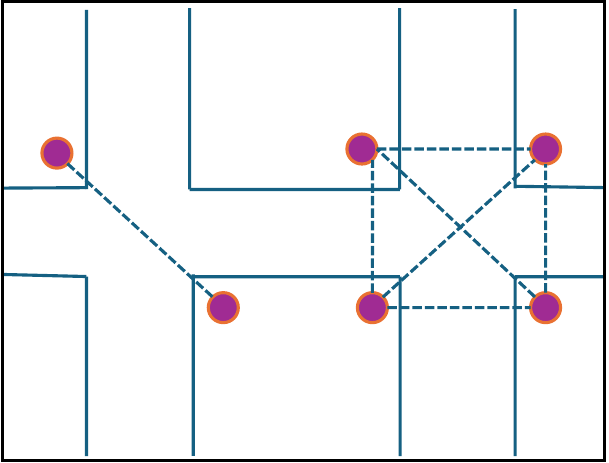}
        \caption{Clusters with full graph connectivity}
        \label{fig:all_clusters}
    \end{subfigure}
    \hfill
    \begin{subfigure}[c]{0.49\textwidth}
        \centering
        \includegraphics[width=0.7\textwidth]{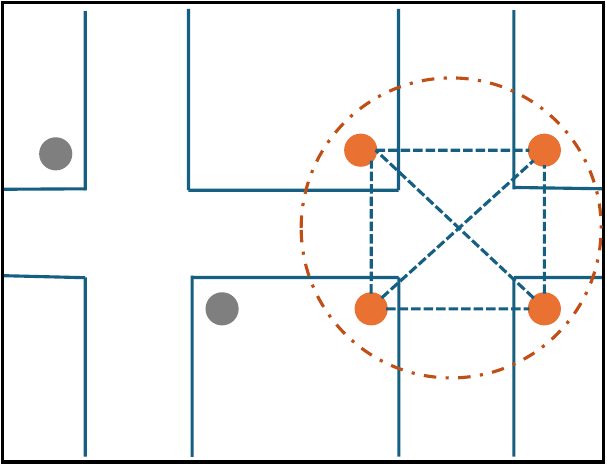}
        \caption{Final cluster with number of signs $\geq$ 3}
        \label{fig:final_cluster}
    \end{subfigure}
    \vspace{0em}
    \begin{center}
        \includegraphics[width=0.8\textwidth]{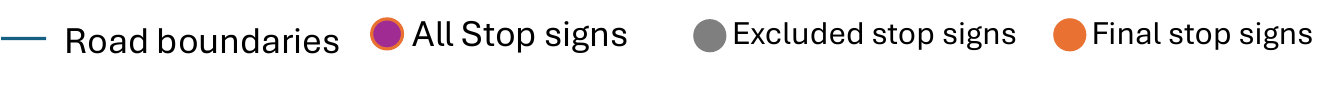}
        \label{fig:clustering_legend}
    \end{center}
    \vspace{-1em}
    \caption{\textcolor{black}{Different steps for the automatic identification of unsignalized intersections with equal priority on all approaches from the raw datasets}}
    \label{fig:clustering}
\end{figure*}

\begin{figure*}[t]
    \centering
    \begin{subfigure}[b]{0.27
    \textwidth}
        \centering
        \includegraphics[width=\textwidth]{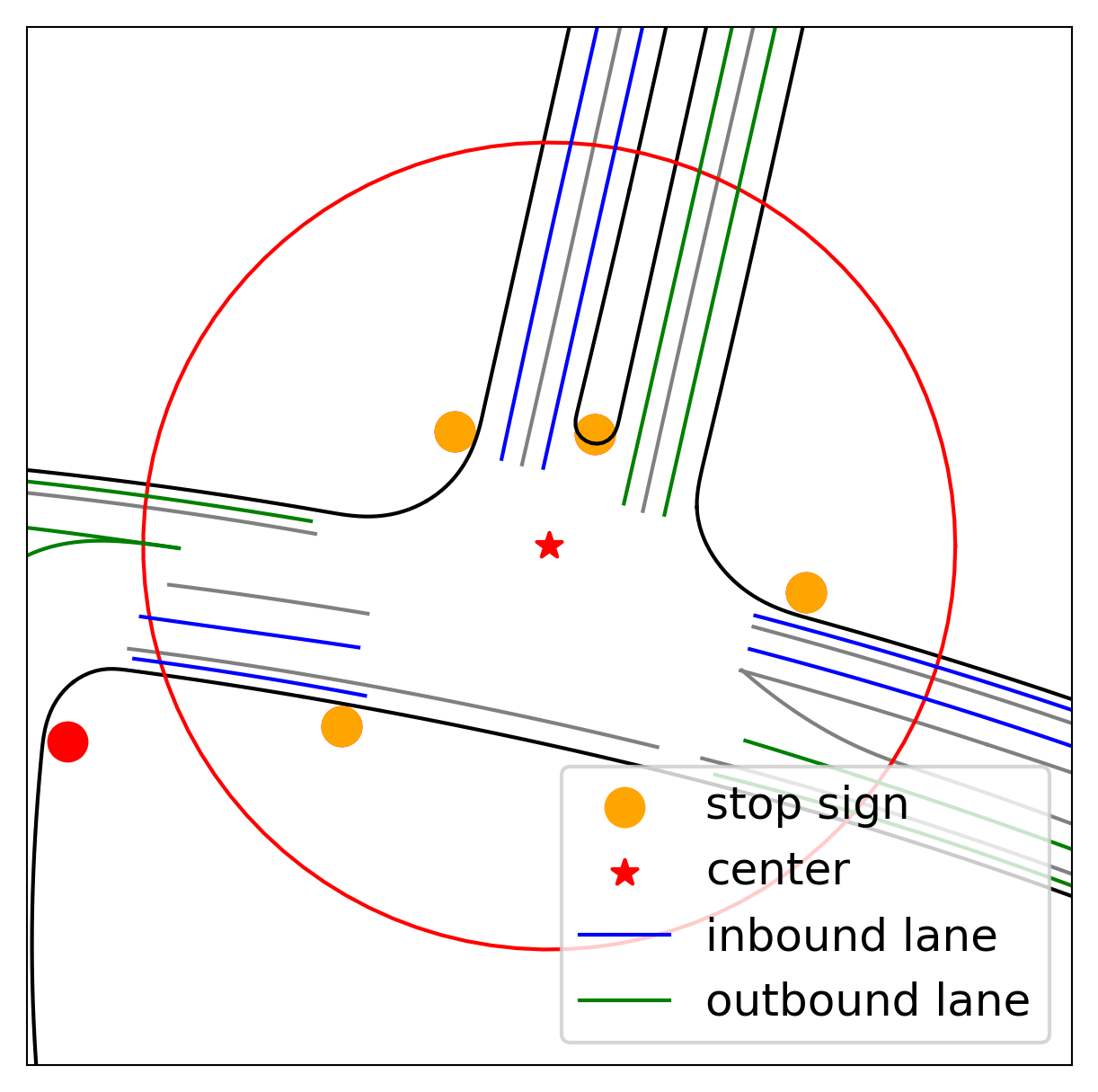}
        \caption{T-intersection}
        \label{fig:example1}
    \end{subfigure}
    \hspace{0.01\textwidth} 
    \begin{subfigure}[b]{0.27\textwidth}
        \centering
        \includegraphics[width=\textwidth]{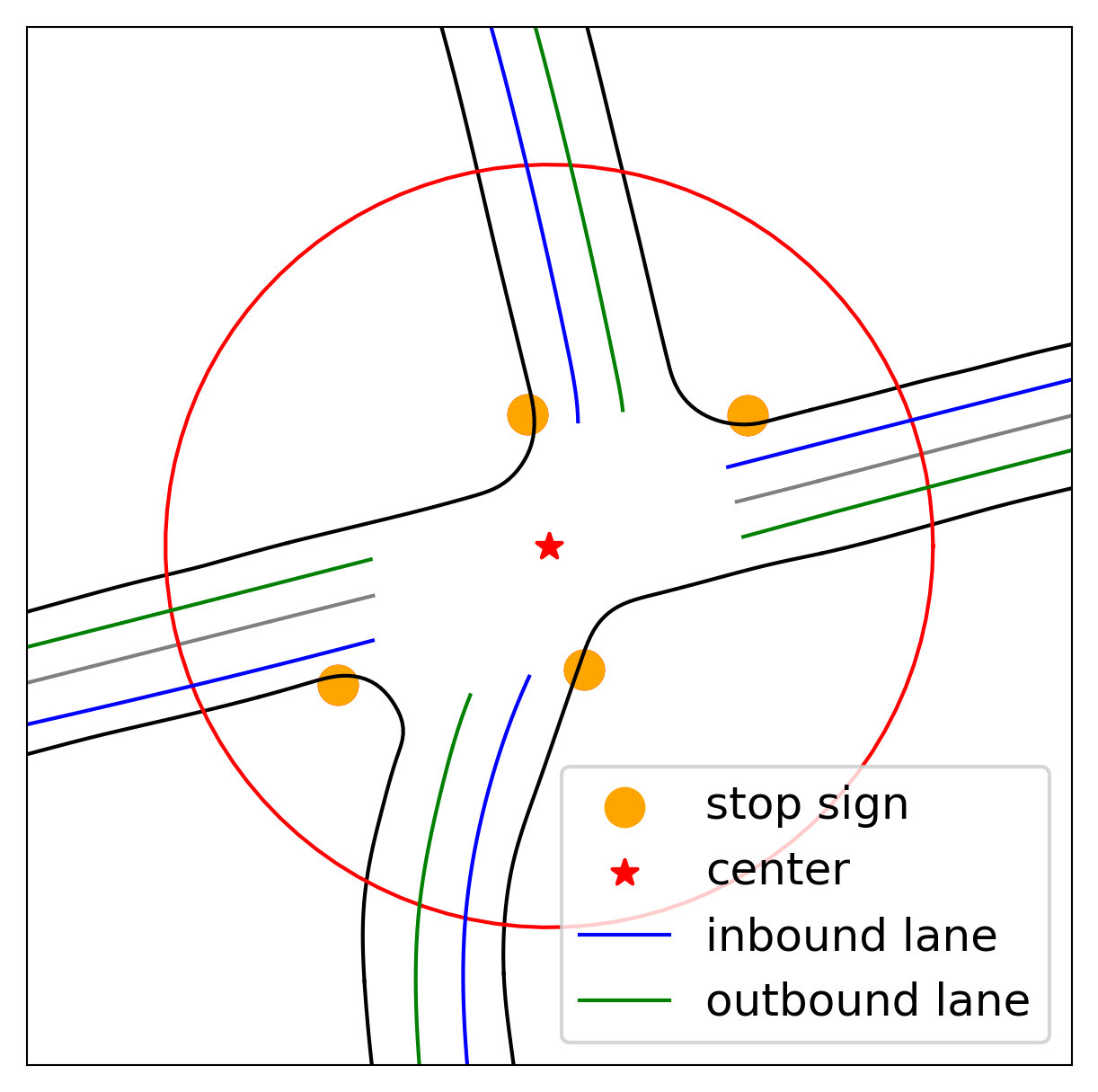}
        \caption{Asymmetric intersection}
        \label{fig:example2}
    \end{subfigure}
    \hspace{0.01\textwidth} 
    \begin{subfigure}[b]{0.27\textwidth}
        \centering
        \includegraphics[width=\textwidth]{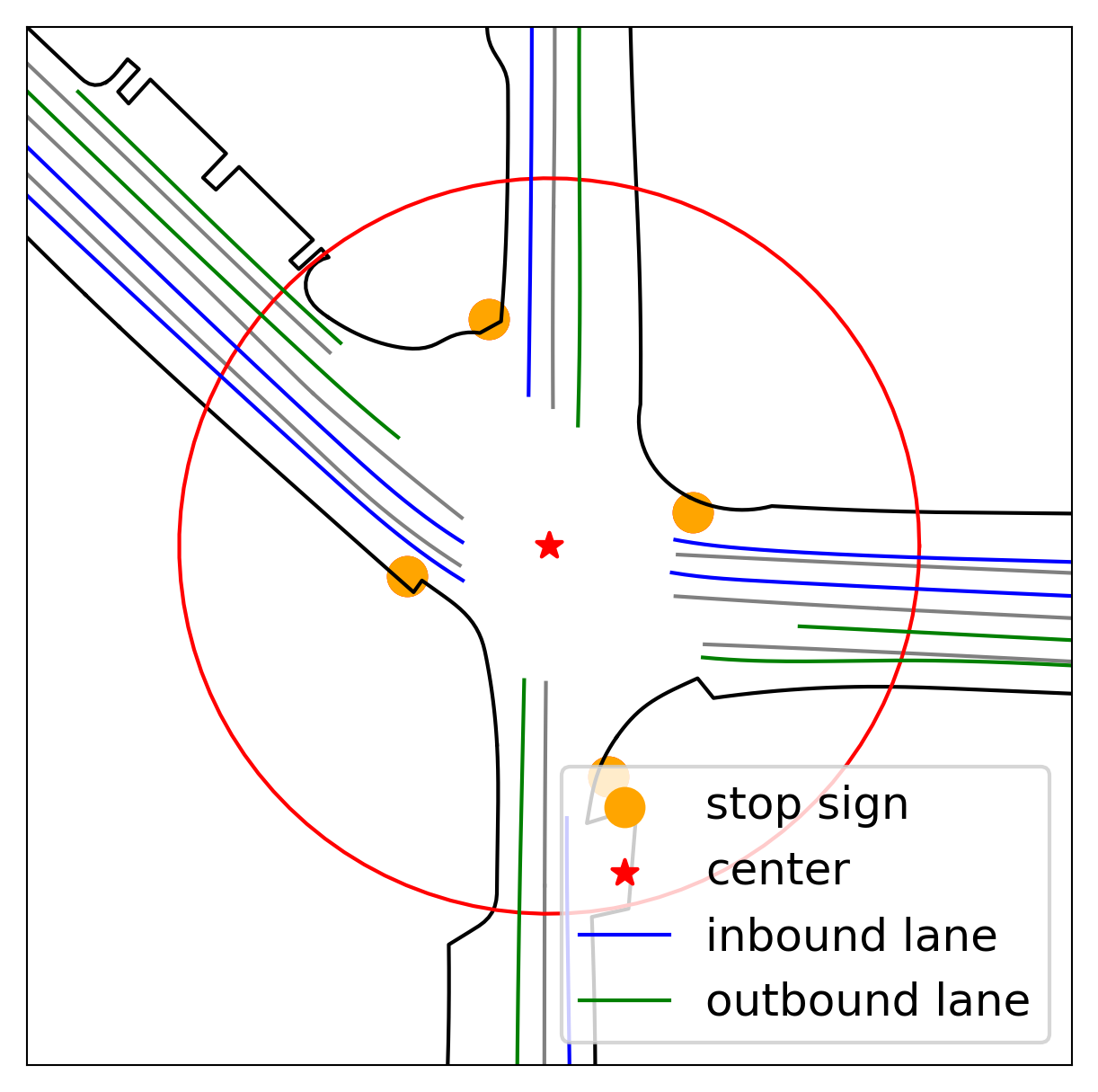}
        \caption{Asymmetric intersection}
        \label{fig:example3}
    \end{subfigure}
    \caption{\textcolor{black}{Examples of identified intersections, highlighting stop signs as indicators of unsignalized intersections with equal priority for all approaches.}}
    \label{fig:intersection_identification}
\end{figure*}

\textcolor{black}{Following data preparation and smoothing, we focus on selecting scenarios that represent unsignalized intersections with equal priority on all approaches. This selection criterion ensures that vehicle interactions are primarily influenced by driving behavior and negotiation processes rather than externally imposed traffic priority rules. To identify these specific scenarios, we implement a three-step approach:}
\paragraph{Extraction of stop signs:} The first step involves identifying the location of stop signs across the road network. To achieve this, we queried the high-definition (HD) map layer and extracted all objects with the semantic label ``stop sign'' (Figure \ref{fig:all_stops}). We chose stop signs as our key indicator because they offer a practical starting point for identifying potential unsignalized intersections. With proper curation and geometric validation, as detailed in the following steps, stop signs can serve as reliable indicators for intersections with equal priority approaches, whereas stop lines or other road markings may also appear at signalized or priority-controlled intersections.
\paragraph{Clustering the stop signs belonging to the same intersection:} In the next step, we identify intersections by grouping stop signs that are close to each other and belong to the same junction. To achieve this, we develop a clustering approach based on maximum clique detection in a proximity graph \cite{wu2015review, bron1973algorithm}. The process begins by using a two-dimensional k-d tree data structure \cite{bentley1975multidimensional, virtanen2020fundamental} to efficiently identify all \textit{pairs} of stop signs that are closer than a specified Euclidean distance threshold (Figure \ref{fig:Pairs}). However, simple pairwise proximity does not guarantee that the entire set forms a compact cluster associated with a \textit{particular} intersection. For example, two stop signs from different intersections might form a pair, while they belong to different junctions. To address this, we construct an $\epsilon$-neighborhood graph \cite{preparata2012computational} where each node represents a stop sign, and edges connect pairs of signs that are within the specified distance threshold. We then apply a maximum clique finding technique based on the Bron–Kerbosch algorithm \cite{bron1973algorithm} to identify the largest fully connected subgraphs where every stop sign is directly connected to every other sign (Figure~\ref{fig:all_clusters}). This approach ensures that all stop signs grouped as part of the same intersection are mutually proximate while preventing the unintended merging of nearby intersections.
\paragraph{Geometric Validation:} While the previous step guarantees identifying individual intersections, additional geometric checks are necessary to confirm that these clusters reflect \textit{all-way stop intersections}. Two criteria are enforced in this validation process: (a) each cluster must contain at least three stop signs to qualify as a valid intersection, including three-leg (T-shaped) layouts; and (b) each approach to the intersection must include at least one stop sign. This second condition ensures the exclusion of priority-controlled intersections. For each cluster that meets these criteria, we compute the intersection center as the centroid of the stop-sign coordinates and define the intersection radius as the maximum distance from this center to any stop sign, plus an additional buffer. This buffer ensures that the defined intersection area sufficiently overlaps with the approaching lanes, enabling accurate detection of vehicles entering or exiting the intersection zone (Figure~\ref{fig:final_cluster}).
The key parameters used in this procedure were determined empirically through iterative testing to balance accurate intersection detection with robustness against false merging. The clustering distance threshold was set to 45 meters, and the intersection buffer to 4 meters. By establishing these key geographical features, we laid the groundwork for the subsequent identification and analysis of merging and crossing conflicts within these intersection areas. Figure~\ref{fig:intersection_identification} shows a few examples of identified intersections, their estimated area, and entering and exiting lanes, together with the stop signs at the end of each approaching lane. These examples demonstrate the capability of the proposed method to correctly identify different types of intersections, from simple intersections to more complex asymmetric ones.

\subsubsection{Conflict Identification and Classification}
\begin{figure}[t]
    \centering
    \begin{subfigure}[b]{0.45\textwidth}
        \centering
        \includegraphics[width=\textwidth]{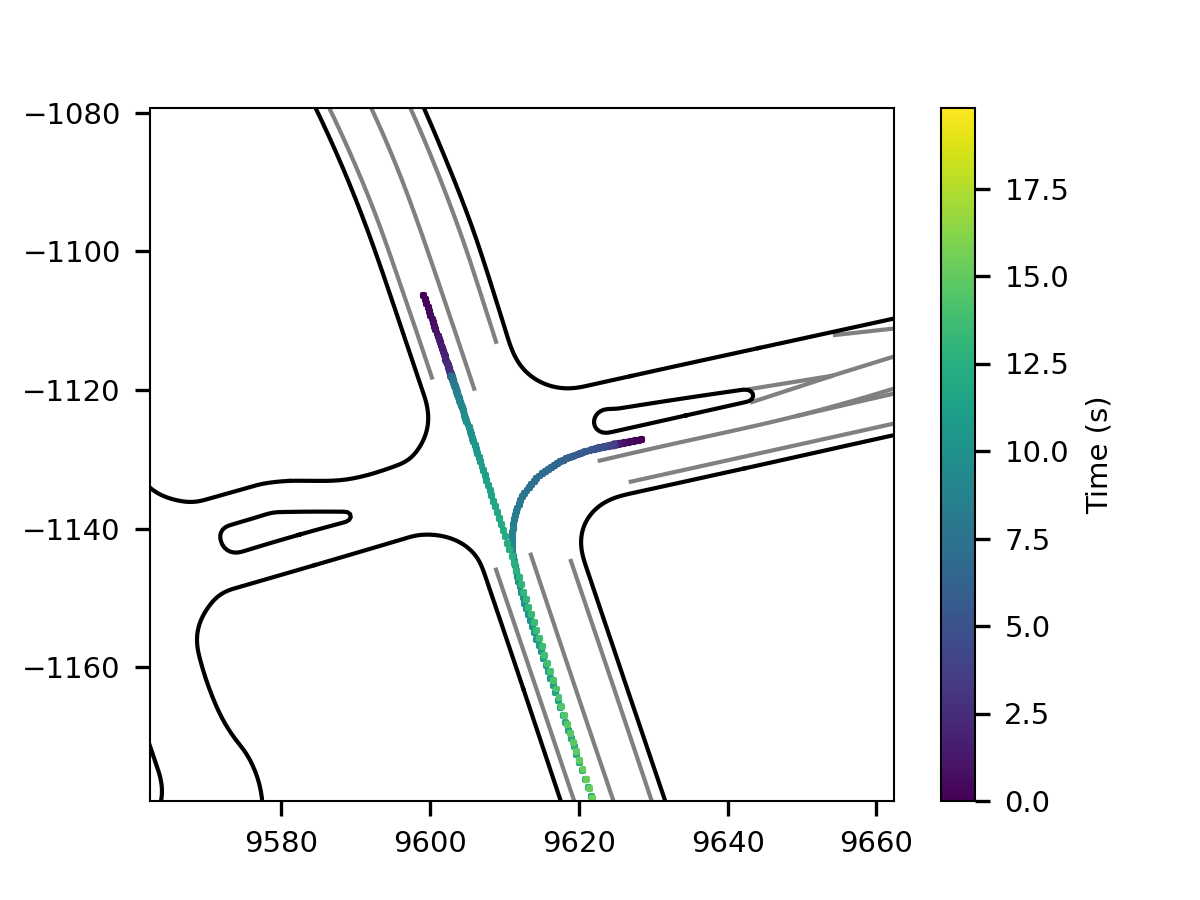}
        \caption{Merging}
        \label{fig:Merging}
    \end{subfigure}
    \begin{subfigure}[b]{0.45\textwidth}
        \centering
        \includegraphics[width=\textwidth]{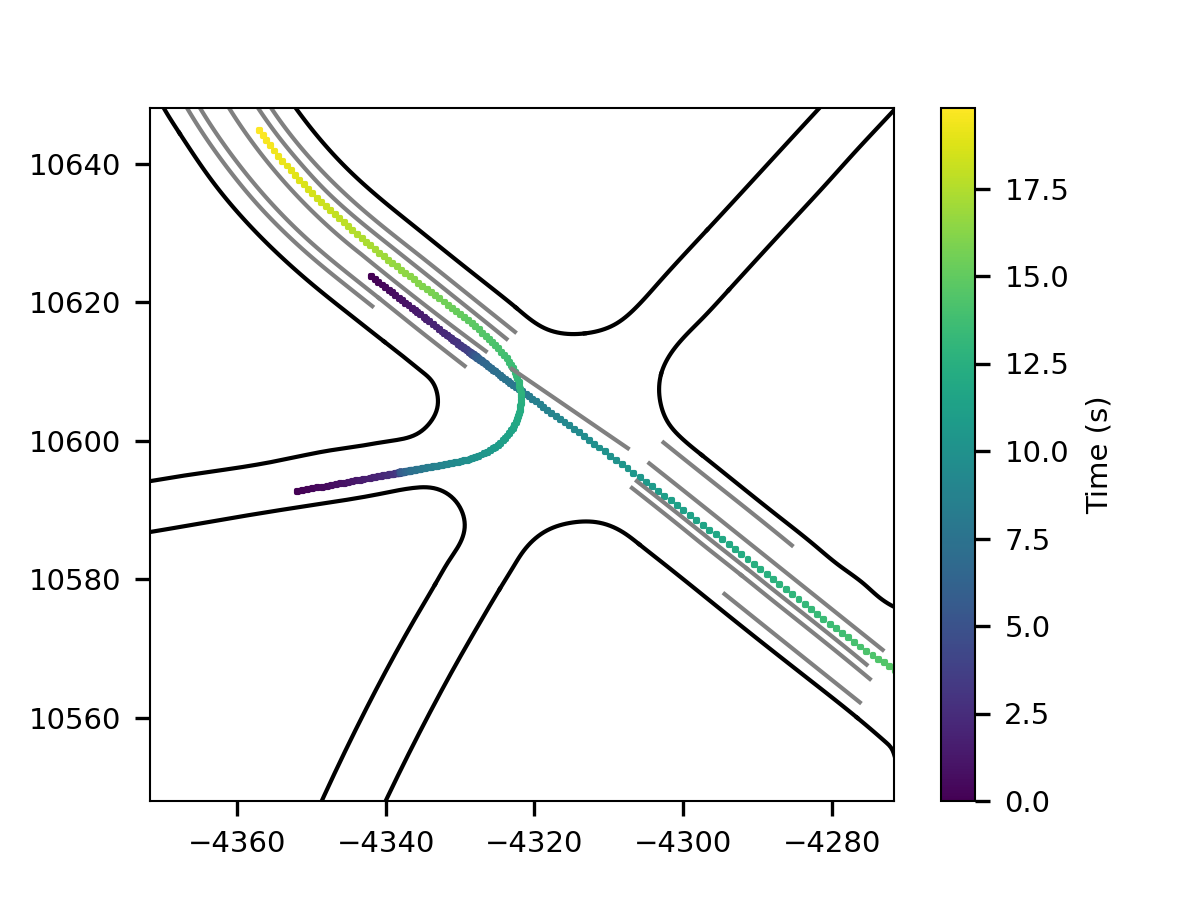}
        \caption{Crossing}
        \label{fig:crossing}
    \end{subfigure}
    \caption{\textcolor{black}{Examples of identified crossing and merging conflicts}}
    \label{fig:merge_cross_example}
\end{figure}

The next step is to identify the merging and crossing conflicts. As a broad definition, a conflict refers to a ``situation where two or more road users approach each other in space and time to such an extent that there is a risk of collision if their movements remain unchanged'' \cite{shahana2023spatiotemporal}. In this study, we take a similar approach proposed by \citep{li2023comparative} with a slight modification. we identify any interaction between two vehicles as a conflict if: 1) the post encroachment time (PET) for that interaction is less than 10 seconds (as opposed to 5 seconds proposed in \cite{li2023comparative}, and the speed of at least one of the vehicles has changed greater than \(3 m/s\). We extended the threshold from 5 seconds to 10 seconds in this study since a significant number of merging interactions, where an automated vehicle passed the conflict point after a human-driven vehicle, exhibited PET values exceeding 5 seconds due to the conservative driving behavior of the automated vehicle. After filtering the desired interactions, they are classified as merging and crossing conflicts using the following criteria:
\begin{itemize}
    \item Merging conflicts: When two vehicles start from different lanes before the intersection and end up in the same lane after the intersection area.
    \item Crossing conflicts: When two vehicles start from different lanes before the intersection and end up in different lanes after the intersection area.
\end{itemize}
Figure~\ref{fig:merge_cross_example} depicts two examples of identified merging and crossing interactions. The color of the points on the trajectory shows the time from the start of the scenario; therefore, the start of the scenario is indicated by a dark black color and the end of the scenario is identified by the light green color. After all, 1424 merging conflicts and 1185 crossing conflicts at unsignalized intersections were identified in the two datasets. Table~\ref{tab:no_scenarios} presents the number of conflicts per dataset and conflict type. In this table, and across the paper, HV-HV refers to conflicts where both interacting vehicles are human-driven, HV-AV refers to the scenarios where the second vehicle in the conflict (follower) is the autonomous vehicle, and AV-HV refers to the cases where the follower is a human-driven vehicle, but the leader is an AV. \textcolor{black}{In this study, the terms ``leader'' and ``follower'' refer to the order of vehicles passing through the conflict zone at unsignalized intersections. The 'leader' is the vehicle that passes first, while the 'follower' is influenced by the leader's decisions and behaviors during the interaction. This terminology is used to describe the dynamics of multi-directional conflicts and does not imply a car-following relationship, as vehicles approach from different directions rather than sequentially in the same lane.}

\begin{table*}[t]
\fontsize{11pt}{12pt}\selectfont
\centering
\caption{Number of scenarios for crossing and merging in Waymo and Lyft datasets}
\begin{tabular}{lcccccccc}
\toprule
 & \multicolumn{4}{c}{Crossing} & \multicolumn{4}{c}{Merging} \\
\cmidrule(lr){2-5} \cmidrule(lr){6-9}
 & Total & HV-HV & HV-AV & AV-HV & Total & HV-HV & HV-AV & AV-HV \\
\midrule
\textbf{Waymo} & 574  & 283 & 142 & 149 & 290  & 107 & 48  & 135 \\
\textbf{Lyft}  & 611  & 456 & 93  & 62  & 1134 & 793 & 190 & 151 \\
\textbf{Total} & 1185 & 739 & 235 & 211 & 1424 & 900 & 238 & 286 \\
\bottomrule
\end{tabular}
\label{tab:no_scenarios}
\end{table*}

\subsubsection{Identifying the Conflict Point}
\begin{figure}[ht]
    \centering
    \begin{subfigure}[b]{0.20\textwidth}
        \centering
        \includegraphics[width=0.9\textwidth]{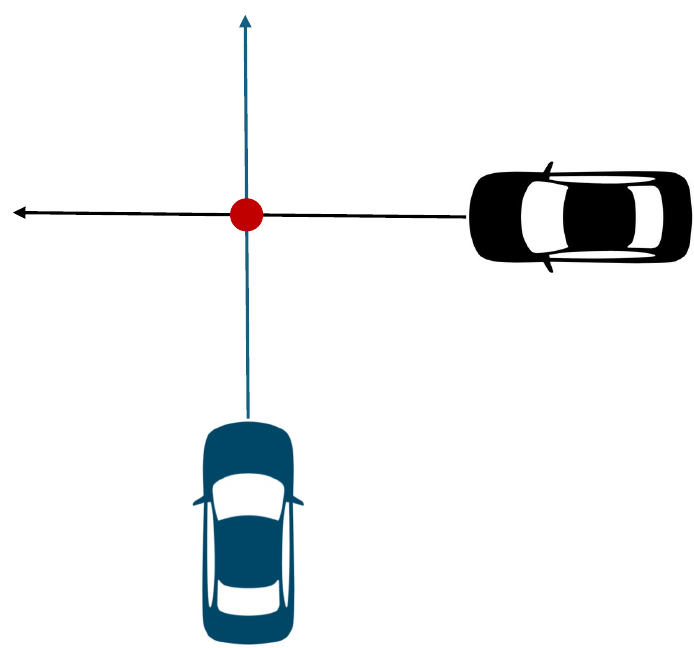}
        \caption{Conflict point for a crossing scenario}
        \label{fig:conflict_crossing}
    \end{subfigure}
    \hspace{0.05\textwidth} 
    \begin{subfigure}[b]{0.2\textwidth}
        \centering
        \includegraphics[width=0.9\textwidth]{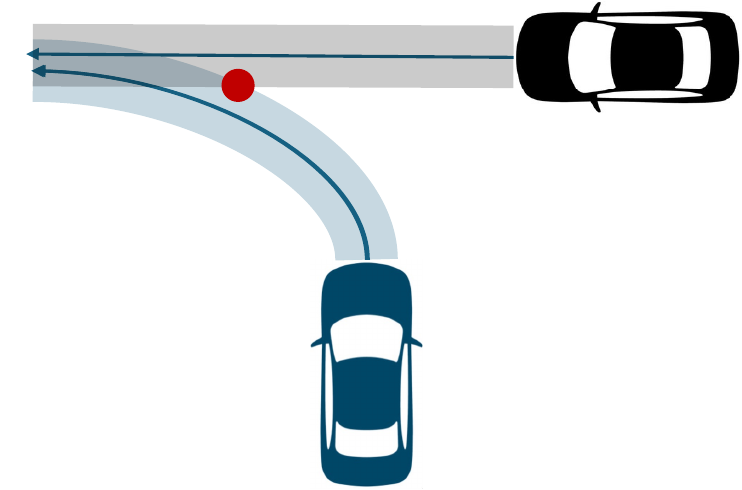}
        \caption{Conflict point for a merging scenario}
        \label{fig:conflict_merging}
    \end{subfigure}
    \caption{Illustration of identifying the conflict point for crossing and merging scenarios.}
    \label{fig:conflict_point}
\end{figure}

For calculating most of the metrics in this study, we first need to identify the potential conflict point for each interaction. For crossing scenarios, this process is straightforward as the conflict point occurs where the trajectories of the two vehicles intersect. However, for merging scenarios, the trajectories of the two vehicles might not intersect due to their lateral offset within the lane. To address this, a buffer of 2 meters (1 meter on each side) is added to each vehicle's trajectory, and the first point of contact between the two buffers is identified as the conflict point. This is shown by the red dot in Figure~\ref{fig:conflict_point}. \textcolor{black}{This buffer size was selected based on typical vehicle widths (1.8-2.5m) \cite{chen2024study} and urban lane widths (3.0-3.7m) \cite{baily2013nacto, hancock2013policy}, ensuring we capture potential conflicts while accounting for lateral positioning variations of vehicles within lanes.}

\textcolor{black}{This structured methodological approach enables a systematic assessment of vehicle interactions at unsignalized intersections. The next section presents the results of this analysis, highlighting key behavioral differences and dynamics between automated and human-driven vehicles.}

\section{Results and Discussions}
In this section, we present the results from our analysis of the behavioral adaptations and differences between HVs and AVs at unsignalized intersections. For our analysis, we classify the interactions into three groups: 
\begin{itemize}
    \item \textit{HV-HV interactions}: These are interactions where both vehicles are human-driven. We analyze these as the baseline to understand typical human driving behaviors at unsignalized intersections. This comparison provides a reference point for evaluating how the introduction of AVs may alter driver behavior and intersection dynamics. 
    \item \textit{AV-HV interactions}: In these interactions, the first vehicle is autonomous, while the second is human-driven. This configuration is critical for examining how AVs may influence the behavior of human drivers who follow them. 
    \item \textit{HV-AV interactions}: Here, the first vehicle is human-driven, and the second is autonomous. These interactions allow us to observe how AVs behave when following human drivers. This setup is especially important for understanding how AVs handle situations where they must yield or adjust to the uncertain behaviors of human drivers, providing insights into the adaptability of AV algorithms in mixed-autonomy traffic. 
\end{itemize}
This classification allows us to systematically explore how AVs and human drivers interact in mixed-autonomy traffic at unsignalized intersections. By isolating these scenarios, we can assess the impact of AVs on driving behaviors, traffic flow, and safety. 

\subsection{Post Encroachment Time and Time-to-Collision}
\begin{figure*}[!ht]
    \centering
    \begin{subfigure}[b]{0.49\textwidth}
        \centering
        \includegraphics[width=\textwidth]{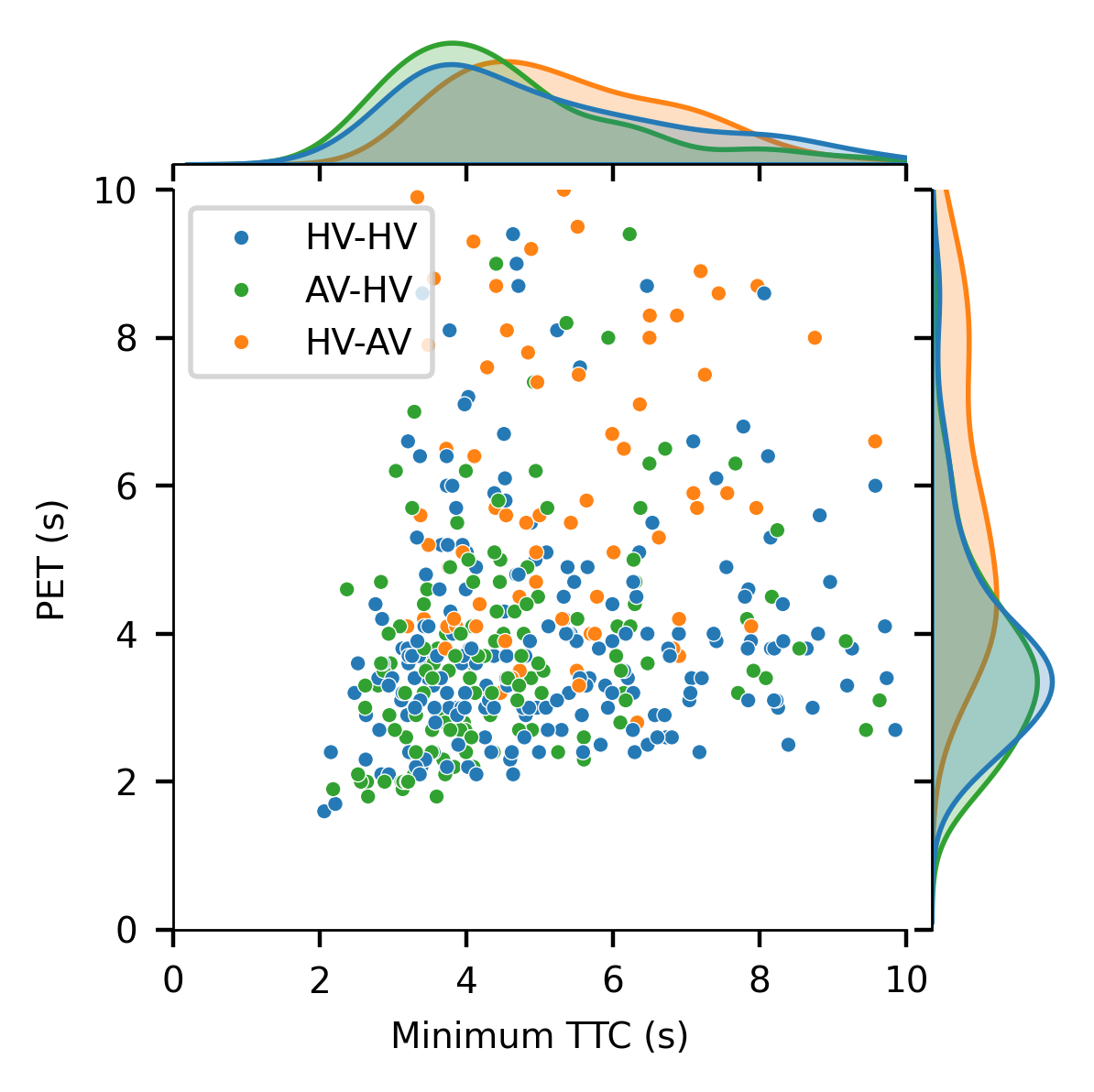}
        \caption{Waymo - Crossing}
        \label{fig:pet_ttc_dist_cross}
    \end{subfigure}
    \hfill
    \begin{subfigure}[b]{0.49\textwidth}
        \centering
        \includegraphics[width=\textwidth]{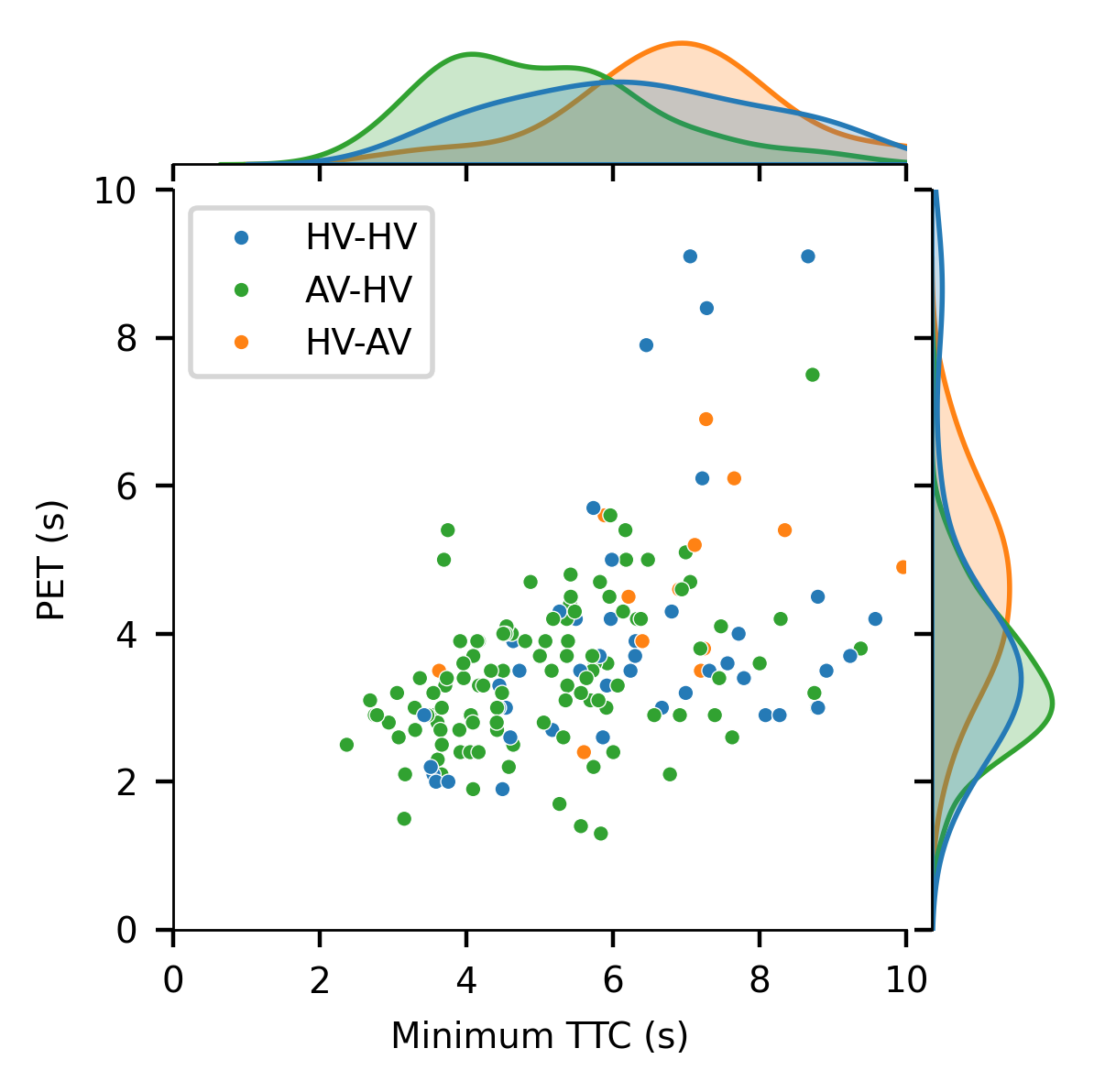}
        \caption{Waymo - Merging}
        \label{fig:pet_ttc_dist_merge}
    \end{subfigure}
    
    \hfill
    \begin{subfigure}[b]{0.49\textwidth}
        \centering
        \includegraphics[width=\textwidth]{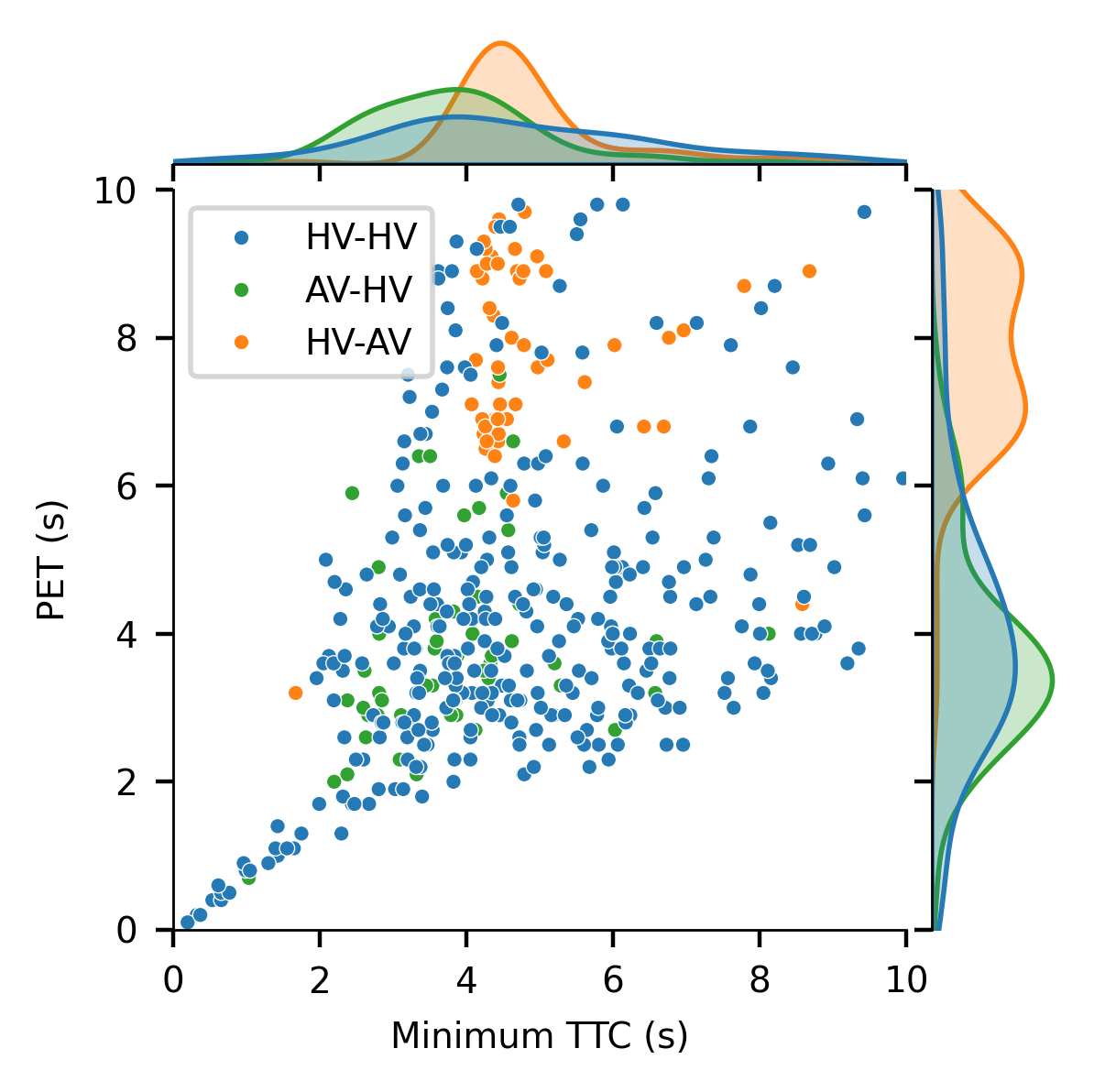}
        \caption{Lyft - Crossing}
        \label{fig:pet_ttc_dist1}
    \end{subfigure}
    \hfill
    \begin{subfigure}[b]{0.49\textwidth}
        \centering
        \includegraphics[width=\textwidth]{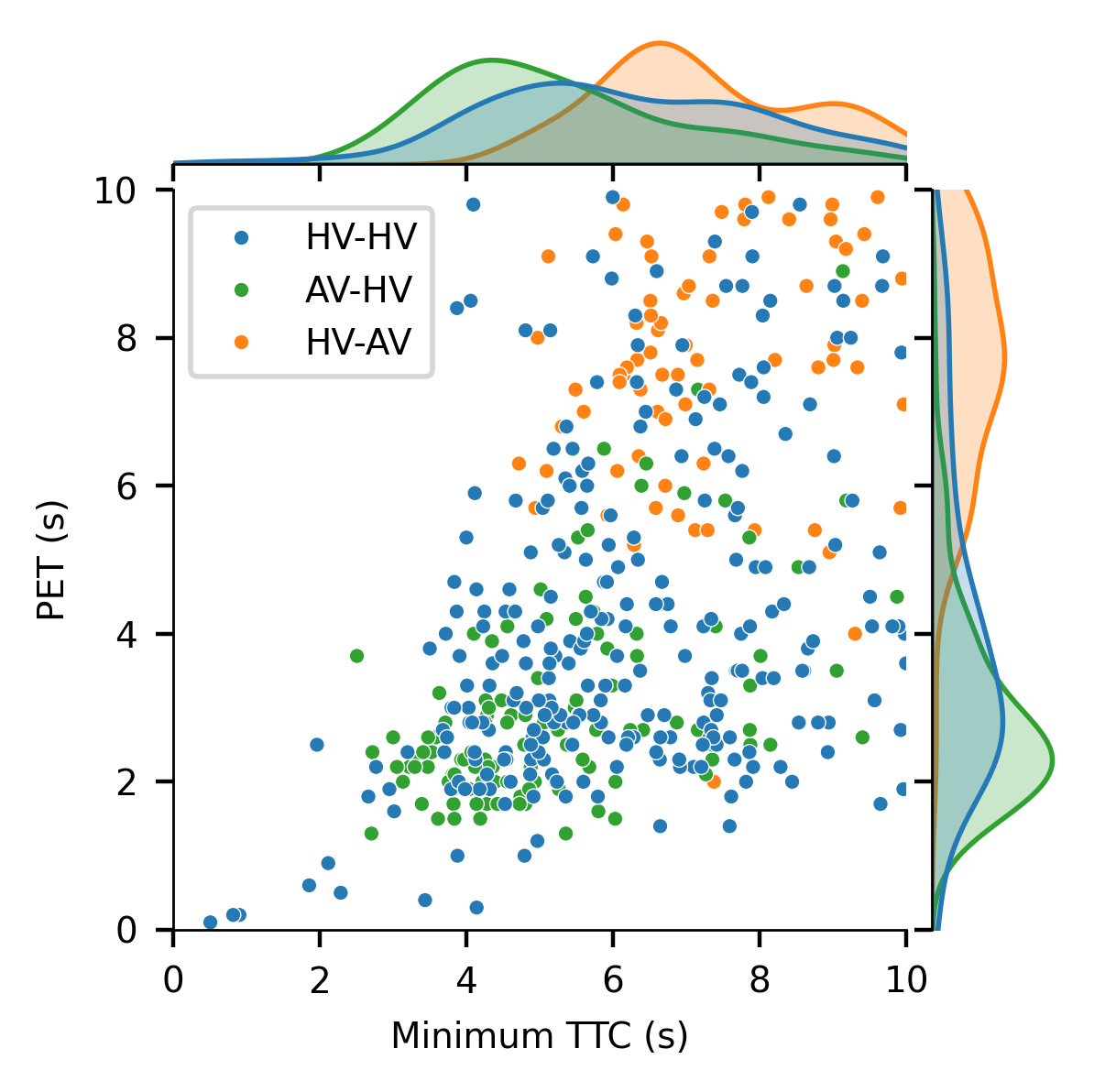}
        \caption{Lyft - Merging}
        \label{fig:pet_ttc_dist2}
    \end{subfigure}
    
    \caption{Joint distributions of PET-minTTC for different types of interactions within Waymo and Lyft datasets}
    \label{fig:pet_ttc_dist}
\end{figure*}

PET and TTC are the two most popular safety surrogate measures that are widely used for the safety analysis of interactions and driving behaviors \cite{wang2021review, ozbay2008derivation}. \textcolor{black}{In this study, we utilized both TTC and PET as complementary safety metrics to provide a comprehensive evaluation of vehicle interactions at unsignalized intersections. PET serves as a retrospective measure, capturing the time gap between two vehicles after the conflict zone is cleared, making it particularly useful for assessing completed maneuvers. TTC, on the other hand, is a proactive metric that evaluates the estimated time remaining to a potential collision based on current speeds and trajectories, allowing for the dynamic assessment of conflict criticality throughout the maneuver.} Figure \ref{fig:pet_ttc_dist} depicts a joint distribution of PET and minimum observed TTC in HV-HV, AV-HV, and HV-AV interactions. \textcolor{black}{We use joint distributions of PET and minimum observed TTC to provide insights into the relationship between these two safety metrics, highlighting how they interact as pre- and post-criticality measures in different interaction types (HV-HV, AV-HV, and HV-AV). Unlike marginal distributions, joint distributions reveal trends, clusters, and correlations that define conflict severity and distinguish crossing from merging interactions.}

The visual inspection of these plots, accompanied by the results of statistical analysis presented in Table \ref{tab:pet_minttc_comparison} reveal interesting findings. In general, in scenarios where an AV follows an HV (orange dots), higher values of PET and minTTC are observed compared to other conflict types. This trend is evident in both merging and crossing scenarios, indicating that AVs maintain larger safety buffers and show more conservative driving behaviors when trailing HVs. These findings highlight the potential of AVs to enhance safety at unsignalized intersections by maintaining larger safety margins. Nevertheless, this benefit comes with a cost of possible lower efficiency due to larger gaps between the vehicles \cite{li2023large}. This is in line with the observations from the Argoverse dataset proposed by \citep{li2023comparative}.

Interestingly, the differences between AV-HV and HV-HV interactions are less apparent, which suggests that human drivers still show similar and relatively aggressive driving styles when interacting with AVs. However, we observed greater heterogeneity in HV behaviors when interacting with other HVs compared to their interactions with AVs. This encourages the argument that the presence of AVs may contribute to reducing variability in HV behavior at intersections, potentially standardizing traffic flow patterns. This finding aligns with observations from car-following scenarios in the literature \cite{wen2022characterizing, zhang2023impact}. However, such a conclusion needs further investigation into a wider range of scenarios and datasets. All in all, the investigation of PET-minTTC metrics indicates that HV-HV and AV-HV conflicts are more prone to observing safety criticality, which highlights the importance of studying the behavior of human drivers in mixed-autonomy traffic. 
\begin{table*}
    \fontsize{11pt}{12pt}\selectfont
    \centering
    \renewcommand{\arraystretch}{1.4}  
    \caption{\textcolor{black}{Comparison of PET and minTTC in crossing and merging scenarios for Waymo and Lyft dataset}}
    \resizebox{\textwidth}{!}{
    \begin{tabular}{ll cccccc cccccc}
    \toprule
    & & \multicolumn{6}{c}{Waymo} & \multicolumn{6}{c}{Lyft} \\
    \cmidrule(lr){3-8} \cmidrule(lr){9-14}
    & & \multicolumn{3}{c}{Crossing} & \multicolumn{3}{c}{Merging} & \multicolumn{3}{c}{Crossing} & \multicolumn{3}{c}{Merging} \\
    \cmidrule(lr){3-5} \cmidrule(lr){6-8} \cmidrule(lr){9-11} \cmidrule(lr){12-14}
    Test & \shortstack{Comparison\\ Benchmark} & $\mu$ ($\sigma$) & t test & u test & $\mu$ ($\sigma$) & t test & u test & $\mu$ ($\sigma$) & t test & u test & $\mu$ ($\sigma$) & t test & u test \\
    \midrule
    \multicolumn{14}{l}{\textbf{PET (s)}} \\
    \midrule
    \textbf{HV-HV} & & 4.08 (1.52) & & & 3.97 (1.88) & & & 4.68 (2.18) & & & 4.62 (2.38) & & \\
    \textbf{AV-HV} & HV-HV & 3.83 (1.46) & 0.10 & 0.08 & 3.43 (0.97) & \textless0.01 & 0.10 & 3.85 (1.28) & \textless0.01 & \textless0.01 & 3.33 (1.89) & \textless0.01 & \textless0.01 \\
    \textbf{HV-AV} & HV-HV & 5.33 (1.74) & \textless0.01 & \textless0.01 & 4.73 (1.69) & 0.02 & \textless0.01 & 7.35 (1.30) & \textless0.01 & \textless0.01 & 7.08 (1.68) & \textless0.01 & \textless0.01 \\
    \textbf{HV-AV} & AV-HV & 5.33 (1.74) & \textless0.01 & \textless0.01 & 4.73 (1.69) & \textless0.01 & \textless0.01 & 7.35 (1.30) & \textless0.01 & \textless0.01 & 7.08 (1.68) & \textless0.01 & \textless0.01 \\
    \midrule
    \multicolumn{14}{l}{\textbf{minTTC (s)}} \\
    \midrule
    \textbf{HV-HV} & & 5.08 (1.84) & & & 6.30 (1.73) & & & 4.71 (1.98) & & & 6.12 (1.89) & & \\
    \textbf{AV-HV} & HV-HV & 4.53 (1.58) & \textless0.01 & \textless0.01 & 5.05 (1.49) & \textless0.01 & \textless0.01 & 3.85 (1.24) & \textless0.01 & \textless0.01 & 5.27 (1.66) & \textless0.01 & \textless0.01 \\
    \textbf{HV-AV} & HV-HV & 5.37 (1.47) & 0.23 & 0.05 & 6.88 (1.49) & 0.28 & 0.26 & 4.81 (1.01) & 0.54 & 0.16 & 7.23 (1.39) & \textless0.01 & \textless0.01 \\
    \textbf{HV-AV} & AV-HV & 5.37 (1.47) & \textless0.01 & \textless0.01 & 6.88 (1.49) & \textless0.01 & \textless0.01 & 4.81 (1.01) & \textless0.01 & \textless0.01 & 7.23 (1.39) & \textless0.01 & \textless0.01 \\
    \bottomrule
    \end{tabular}
    }
    \label{tab:pet_minttc_comparison}
\end{table*}

\subsection{Maximum Required Deceleration (MRD)}\label{sec:mrd}
\begin{figure}[t]
    \centering
    \begin{subfigure}[b]{0.23\textwidth}
        \centering
        \includegraphics[width=\textwidth]{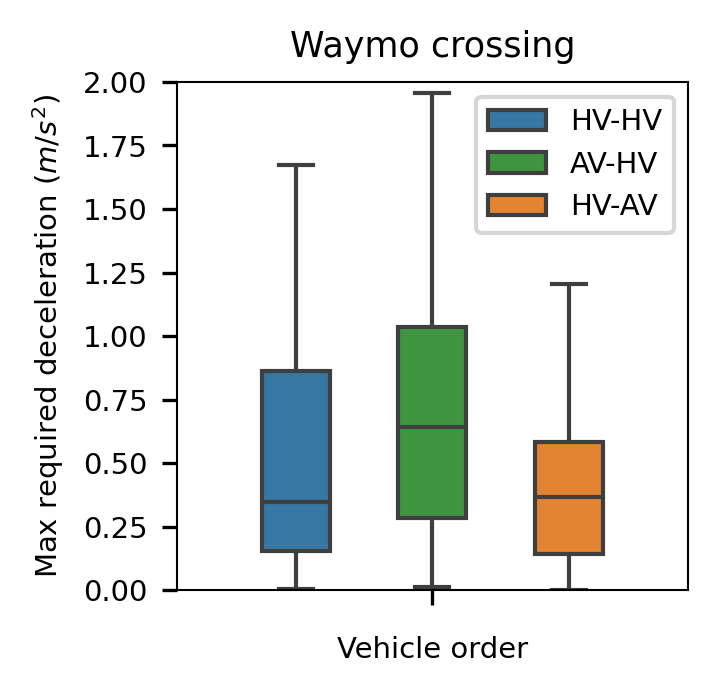}
        \caption{Waymo - Crossing}
    \end{subfigure}
    \begin{subfigure}[b]{0.23\textwidth}
        \centering
        \includegraphics[width=\textwidth]{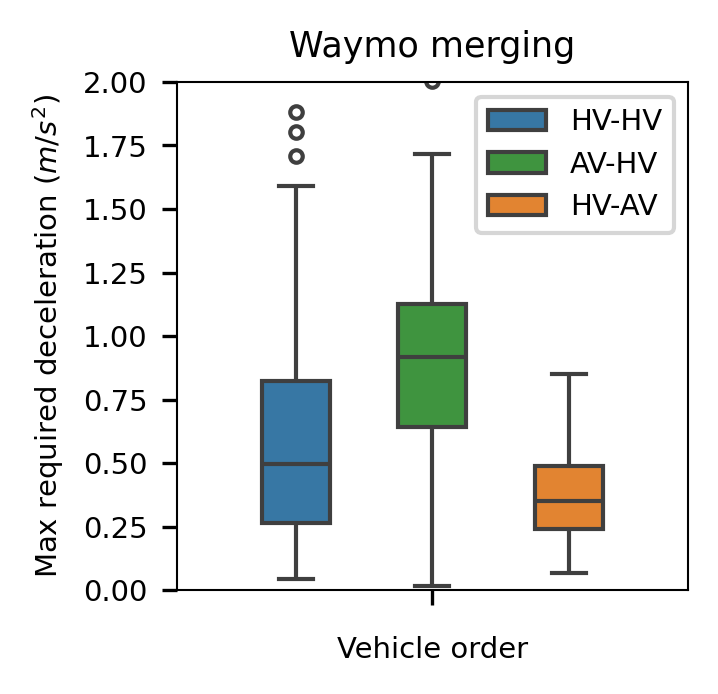}
        \caption{Waymo - Merging}
    \end{subfigure}
    \begin{subfigure}[b]{0.23\textwidth}
        \centering
        \includegraphics[width=\textwidth]{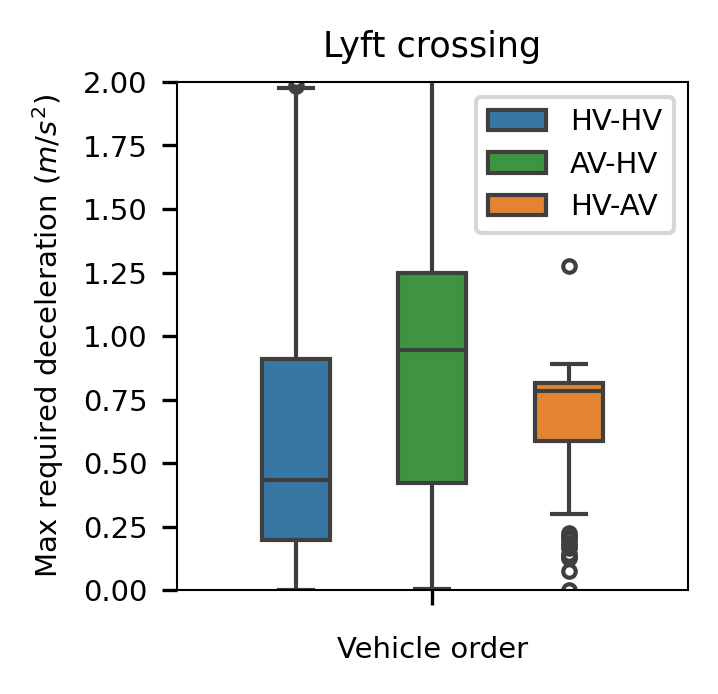}
        \caption{Lyft - Crossing}
    \end{subfigure}
    \begin{subfigure}[b]{0.23\textwidth}
        \centering
        \includegraphics[width=\textwidth]{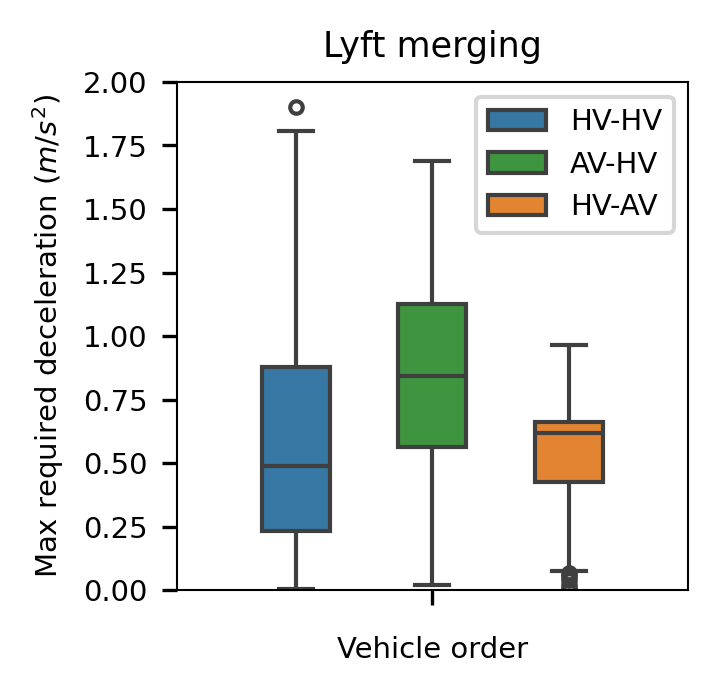}
        \caption{Lyft - Merging}
    \end{subfigure}

    \caption{comparison of MRD values for Waymo and Lyft vehicles in merging and crossing scenarios}
    \label{fig:mrd}
\end{figure}

We employ maximum required deceleration (MRD) as another indication of scenario criticality. Higher values of MRD indicate a harsher possible reaction from the follower considering its current speed and distance to the leader in order to avoid a potential collision. Figure \ref{fig:mrd} depicts the boxplots of estimated MRD values for the followers before reaching the conflict point, and Table \ref{tab:mrd_comparison} presents the results of statistical tests for evaluating the significance of observed differences. Interestingly, AV-HV interactions show higher values of MRD compared to other interaction types for both merging and crossing scenarios. This might be related to the unexpected behavior of AVs, which can lead to misunderstandings by human drivers. Specifically, human drivers may be uncertain whether the AV is yielding or proceeding, causing hesitation and delayed deceleration. This uncertainty can result in human drivers delaying their deceleration, leading to higher MRD values. This observation aligns with reports on crashes involving automated vehicles, where the unexpected behavior of AVs has been identified as a key factor in AV-HV collisions, particularly in rear-end collisions in car-following scenarios \cite{boggs2020exploratory}. Our findings at unsignalized intersections suggest that similar issues could arise in more complex traffic scenarios as AVs become more prevalent. Moreover, the high dispersion of MRD values in HV-HV and AV-HV scenarios compared to HV-AV scenarios emphasizes the heterogeneity of the following behavior of HVs compared to AVs. This has been reported in studies related to car-following scenarios as well \cite{zhao2020field, ma2024driver}. 

\textcolor{black}{Comparing the MRD values between the Waymo and Lyft datasets reveals a key distinction. Lyft vehicles, when following HVs, exhibit higher MRD values, suggesting that human drivers are more likely to force Lyft vehicles to yield, often pushing them to decelerate more harshly. This could lead to potentially unsafe situations. In contrast, Waymo vehicles show significantly lower MRD values in HV-AV interactions (their average is close to HV-HV interactions, with fewer variations), indicating they are less often bullied by human drivers, likely due to their more human-compatible driving style. This observation provides two important insights: First, it highlights the variability in AVs' behavior across different manufacturers and its potential impact on traffic safety and dynamics, and second, it underscores the importance of human-predictable decision-making in AVs for their safe and effective integration into mixed-autonomy traffic environments.}

All in all, the MRD analysis, along with PET and TTC observations, indicates that while automated vehicles generally behave more safely and conservatively than human drivers, their decisions and actions may not fully align with human expectations. This misalignment can cause confusion for human drivers, potentially leading to unsafe situations. Furthermore, the overly cautious behavior of AVs can inadvertently provoke aggressive driving styles in human drivers. These findings highlight the need for AVs to adopt behavior that is not only safe but also predictable to human drivers to ensure smooth and safe integration into traffic.

\begin{table*}[t]
\fontsize{11pt}{12pt}\selectfont
\centering
\caption{Comparison of Maximum Required Deceleration (MRD) for Waymo and Lyft datasets in crossing and merging scenarios}
\renewcommand{\arraystretch}{1.4}  
\resizebox{\textwidth}{!}{
\begin{tabular}{ll cccccc cccccc}
\toprule
& & \multicolumn{6}{c}{Waymo} & \multicolumn{6}{c}{Lyft} \\
\cmidrule(lr){3-8} \cmidrule(lr){9-14}
& & \multicolumn{3}{c}{Crossing} & \multicolumn{3}{c}{Merging} & \multicolumn{3}{c}{Crossing} & \multicolumn{3}{c}{Merging} \\
\cmidrule(lr){3-5} \cmidrule(lr){6-8} \cmidrule(lr){9-11} \cmidrule(lr){12-14}
Test & Benchmark & $\mu$ ($\sigma$) & t test & u test & $\mu$ ($\sigma$) & t test & u test & $\mu$ ($\sigma$) & t test & u test & $\mu$ ($\sigma$) & t test & u test \\
\midrule
\multicolumn{14}{l}{\textbf{MRD ($\mathbf{\text{m/s}^2}$)}} \\
\midrule
\textbf{HV-HV} & & 0.53 (0.47) & & & 0.58 (0.41) & & & 0.61 (0.52) & & & 0.59 (0.49) & & \\
\textbf{AV-HV} & HV-HV & 0.69 (0.47) & \textless0.01 & \textless0.01 & 0.89 (0.43) & 0.07 & \textless0.01 & 0.88 (0.56) & \textless0.01 & \textless0.01 & 0.85 (0.39) & \textless0.01 & \textless0.01\\
\textbf{HV-AV} & HV-HV & 0.39 (0.28) & \textless0.01 & 0.05 & 0.38 (0.18) & \textless0.01 & \textless0.01 & 0.65 (0.25) & 0.38 & \textless0.01 & 0.54 (0.23) & 0.15 & 0.75 \\
\textbf{HV-AV} & AV-HV & 0.39 (0.28) & \textless0.01 & \textless0.01 & 0.38 (0.18) & \textless0.01 & \textless0.01 & 0.65 (0.25) & \textless0.01 & \textless0.01 & 0.54 (0.23) & \textless0.01 & \textless0.01 \\
\bottomrule
\end{tabular}
}
\label{tab:mrd_comparison}
\end{table*}


\subsection{Time Advantage Distributions}
In this study, we propose a novel approach to identify potentially aggressive driving behavior by the ``leading vehicle'' at intersections. To this end, we utilize the metric called "time advantage" (TA) to estimate which vehicle will reach the intersection first at each time step from when the conflict is detected until the first vehicle finally passes the conflict point. Tracking the TA values as the vehicles approach the intersection allows us to determine which vehicle established a positional advantage during the interaction. For instance, observing frequent negative TA values for the leading vehicle at an interaction suggests that the lead vehicle was initially at a disadvantage (negative TA values) but may have accelerated or acted aggressively to gain the lead at the conflict point. Therefore, TA distributions allow us to quantify aggressive maneuvers, such as not decelerating or forcing the other vehicle to yield. \textcolor{black}{It is important to note that while HV-AV and AV-HV classifications in this study refer to the ``final passing'' order of vehicles at the intersection, the TA values represent the ``estimated'' passing order at each time instance during the \textit{approaching phase}. Therefore, negative TA values for an AV-HV interaction indicate moments when the vehicle that ultimately passed first (AV) was temporarily projected to arrive second based on instantaneous speeds and distances (and therefore had negative TA values for those time steps). These variations in TA values during the approaching phase help identify how vehicles negotiate priority and adjust their behavior before reaching the conflict point.}

\begin{figure}[t]
    \centering
    \begin{subfigure}[b]{0.23\textwidth}
        \centering
        \includegraphics[width=\textwidth]{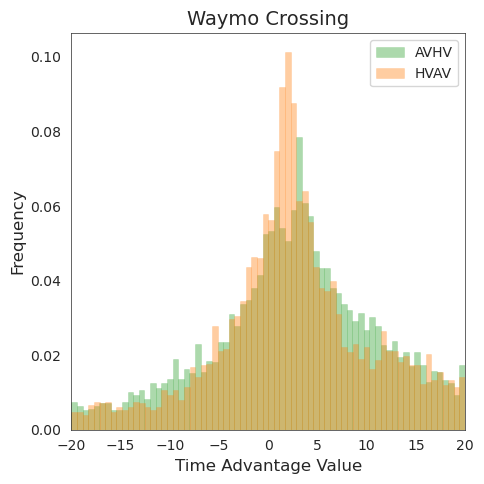}
        \caption{Waymo - Crossing}
    \end{subfigure}
    \begin{subfigure}[b]{0.23\textwidth}
        \centering
        \includegraphics[width=\textwidth]{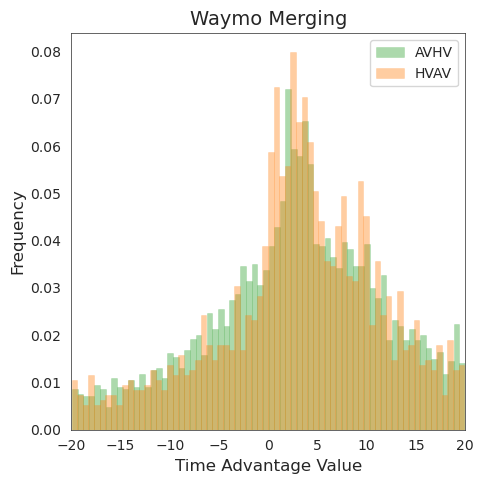}
        \caption{Waymo - Merging}
    \end{subfigure}
    \begin{subfigure}[b]{0.23\textwidth}
        \centering
        \includegraphics[width=\textwidth]{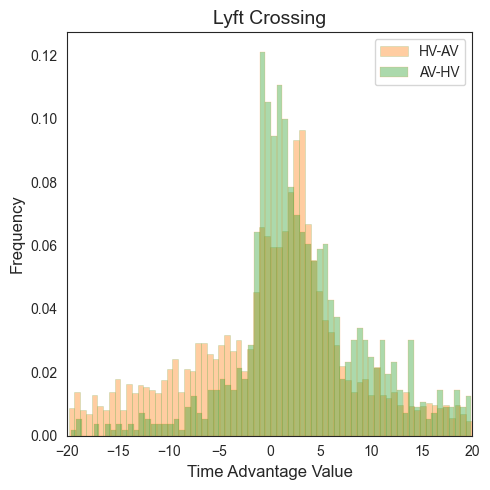}
        \caption{Lyft - Crossing}
    \end{subfigure}
    \begin{subfigure}[b]{0.23\textwidth}
        \centering
        \includegraphics[width=\textwidth]{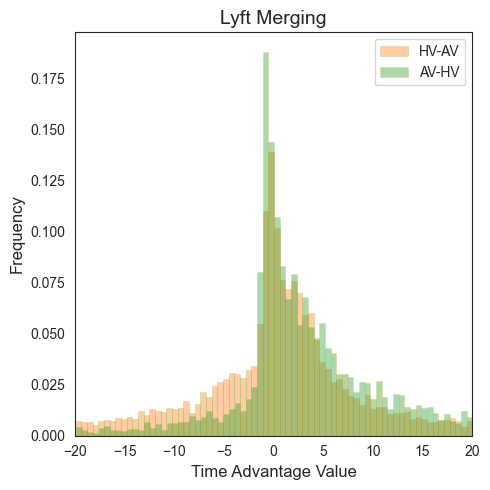}
        \caption{Lyft - Merging}
    \end{subfigure}

    \caption{TA value distributions for Waymo and Lyft datasets}
    \label{fig:ta}
\end{figure}

The TA distributions for HV-AV and AV-HV interactions for both Waymo and Lyft vehicles are depicted in Figure \ref{fig:ta}. The statistical tests for comparing these distributions are provided in Table \ref{tab:ta_test}.
\textcolor{black}{To compare the TA distributions between different interaction types (HV-AV vs. AV-HV), we employ two-sample Kolmogorov-Smirnov (KS) and two-sample Anderson-Darling (AD) tests. These two tests play a complementary role to each other. The KS test evaluates the maximum distance between two cumulative distribution functions and is particularly sensitive to differences around the median of the distributions. The two-sample AD test, while similar in purpose, gives more weight to observations in the tails of the distributions and generally provides higher statistical power for detecting distributional differences \cite{engmann2011comparing}. These tests are specifically chosen (for example, over the Chi-Square test) for their ability to compare continuous distributions without requiring data binning, which would result in loss of information and precision.}

These analyses reveal interesting patterns: The distribution of TA values for Waymo vehicles when being the final follower in an interaction closely resembles that of human drivers in comparable situations. This suggests that Waymo vehicles show human-comparable behaviors during the negotiation phase when approaching the intersection. In contrast, the distribution of TA values for Lyft vehicles differs significantly from human drivers (Table \ref{tab:ta_test}). Visual inspection of the distributions suggests that human drivers tend to take advantage of Lyft vehicles more often than Waymo vehicles in both merging and crossing conflicts (as shown by the more frequent occurrence of negative TA values when the human driver ultimately becomes the leader in the interaction (i.e., HV-AV interactions)). This could be due to Lyft's more conservative driving style. These observations about the behavior of Waymo and Lyft vehicles are in line with previous observations in Section \ref{sec:mrd}, where Lyft vehicles often exhibited higher MRD values, indicating a need for harder braking to avoid collisions, likely due to more aggressive driving from human drivers.

\begin{table*}
\fontsize{10pt}{11pt}\selectfont
    \centering
    \caption{\textcolor{black}{Statistical test results of comparing the TA value distributions of AV-HV and HV-AV interactions for Waymo and Lyft datasets}}
    \label{tab:ta_test}
    \begin{tabularx}{\textwidth}{lXXXXXXXX}
        \toprule
        & \multicolumn{4}{c}{Waymo} & \multicolumn{4}{c}{Lyft} \\
        \cmidrule(lr){2-5} \cmidrule(lr){6-9}
        & \multicolumn{2}{c}{Crossing} & \multicolumn{2}{c}{Merging} & \multicolumn{2}{c}{Crossing} & \multicolumn{2}{c}{Merging} \\
        \cmidrule(lr){2-3} \cmidrule(lr){4-5} \cmidrule(lr){6-7} \cmidrule(lr){8-9}
        Test & Statistic & p-value & Statistic & p-value & Statistic & p-value & Statistic & p-value \\
        \midrule
        \textbf{\textcolor{black}{two-sample}} \vspace{2pt} \\ \textbf{Kolmogorov-Smirnov} & 0.133 & 0.825 & 0.147 & 0.863 & 0.184 & 0.000 & 0.250 & 0.000 \\
        \midrule
        \textbf{\textcolor{black}{two-sample}} \vspace{2pt} \\ \textbf{Anderson-Darling} & -0.179 & 0.250 & -0.701 & 0.250 & 173.958 & 0.001 & 500.884 & 0.001 \\
        \bottomrule
    \end{tabularx}
\end{table*}

\subsection{Speed and Acceleration Analysis}
The speed of vehicles within the intersection and their acceleration behavior are important aspects that can be used both for studying the safety and efficiency of the intersection and for the calibration of microscopic traffic flow models. Our analysis of these metrics focuses on two aspects: the follower's speed at the conflict point, which serves as an indicator of the aggressiveness of the following vehicle in an interaction; and the speed and acceleration profiles when the vehicle starts from a standstill, which can influence traffic efficiency and performance.

\begin{figure}
    \centering
    \begin{subfigure}[b]{0.23\textwidth}
        \centering
        \includegraphics[width=\textwidth]{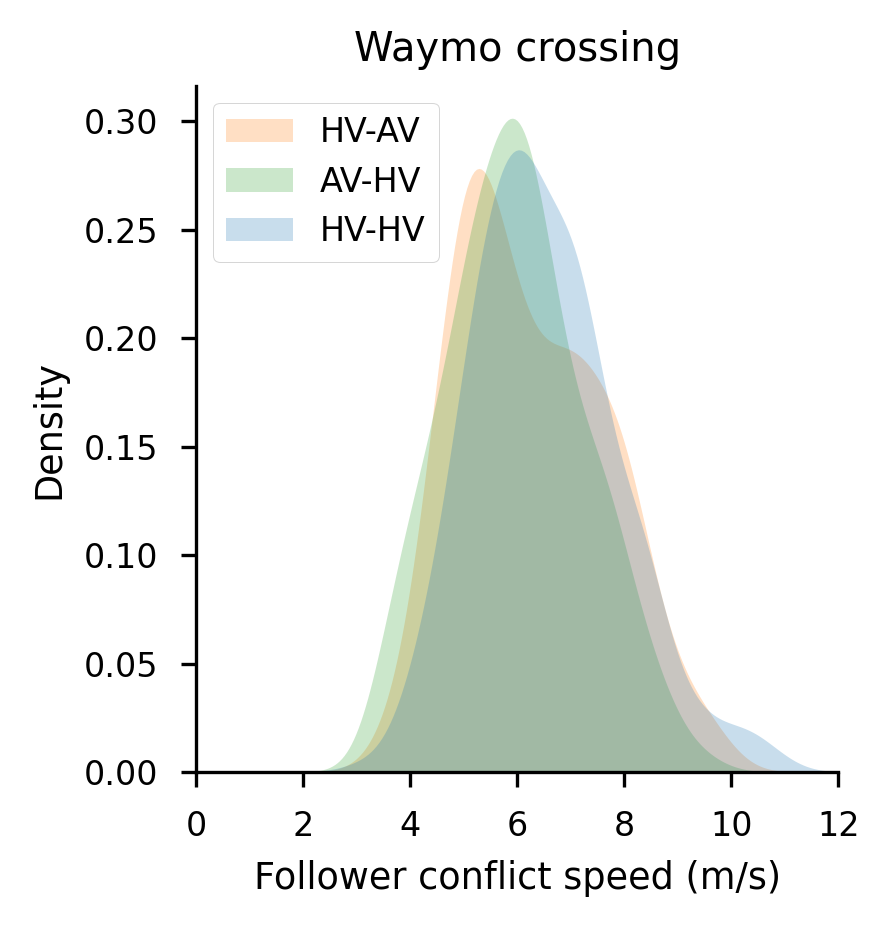}
        \caption{Waymo - Crossing}
    \end{subfigure}
    \begin{subfigure}[b]{0.23\textwidth}
        \centering
        \includegraphics[width=\textwidth]{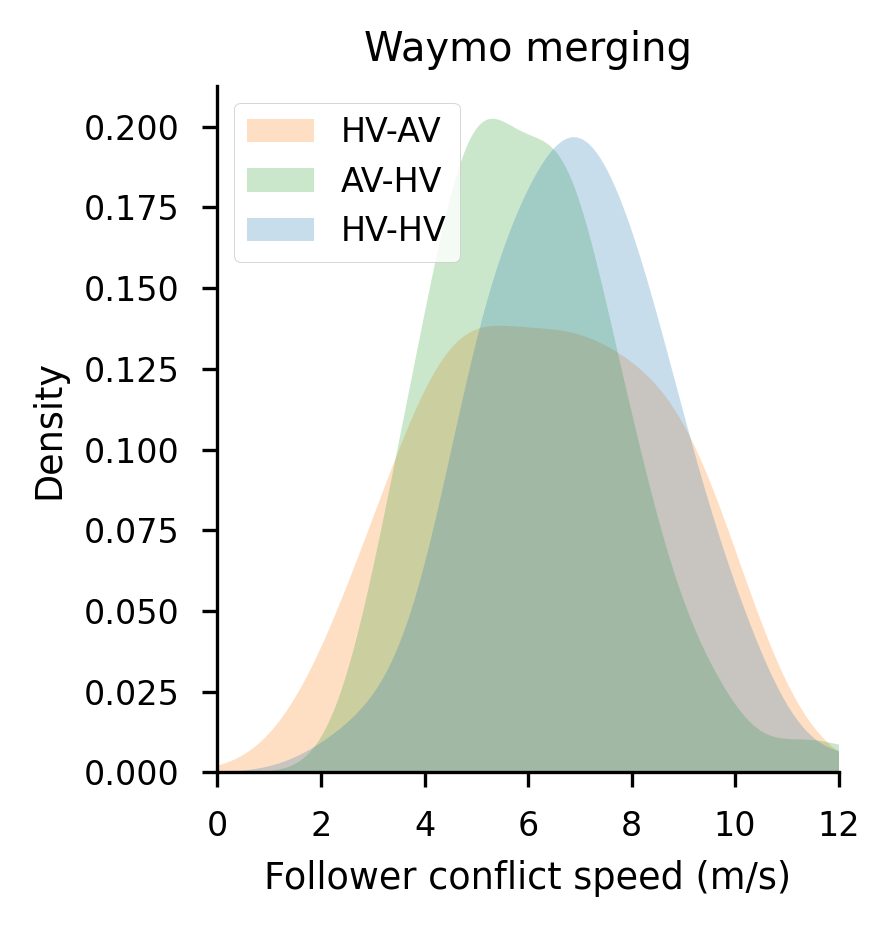}
        \caption{Waymo - Merging}
    \end{subfigure}
    \begin{subfigure}[b]{0.23\textwidth}
        \centering
        \includegraphics[width=\textwidth]{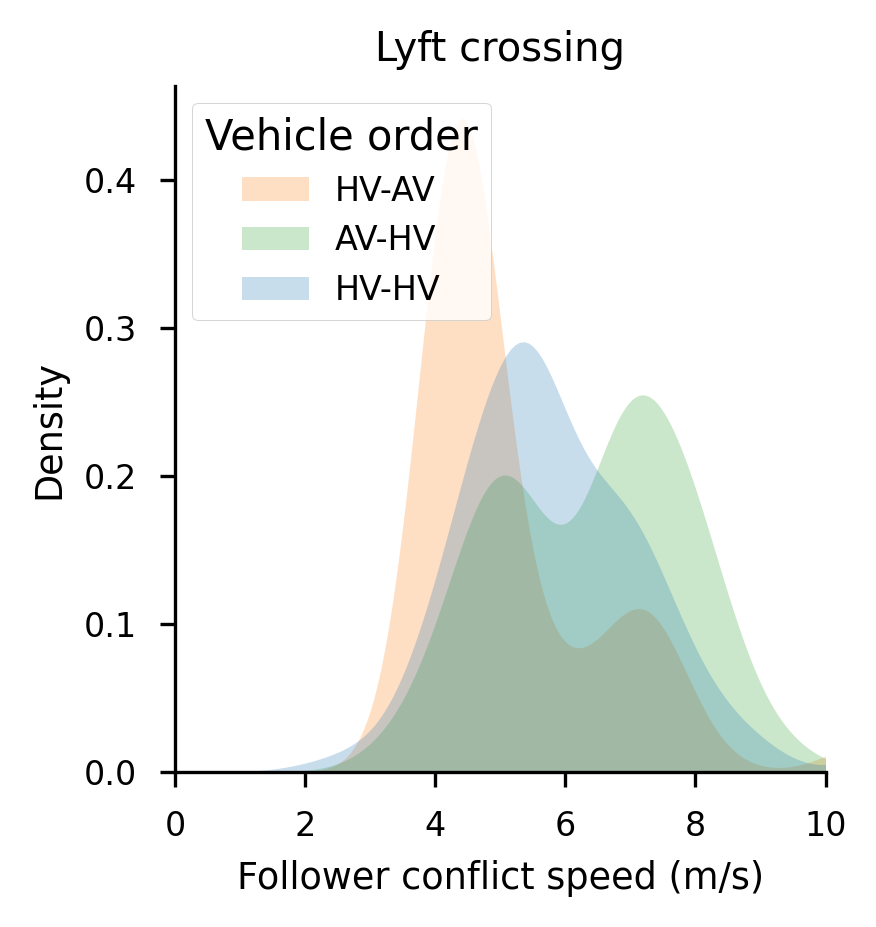}
        \caption{Lyft - Crossing}
        \label{fig:sub3}
    \end{subfigure}
    \begin{subfigure}[b]{0.23\textwidth}
        \centering
        \includegraphics[width=\textwidth]{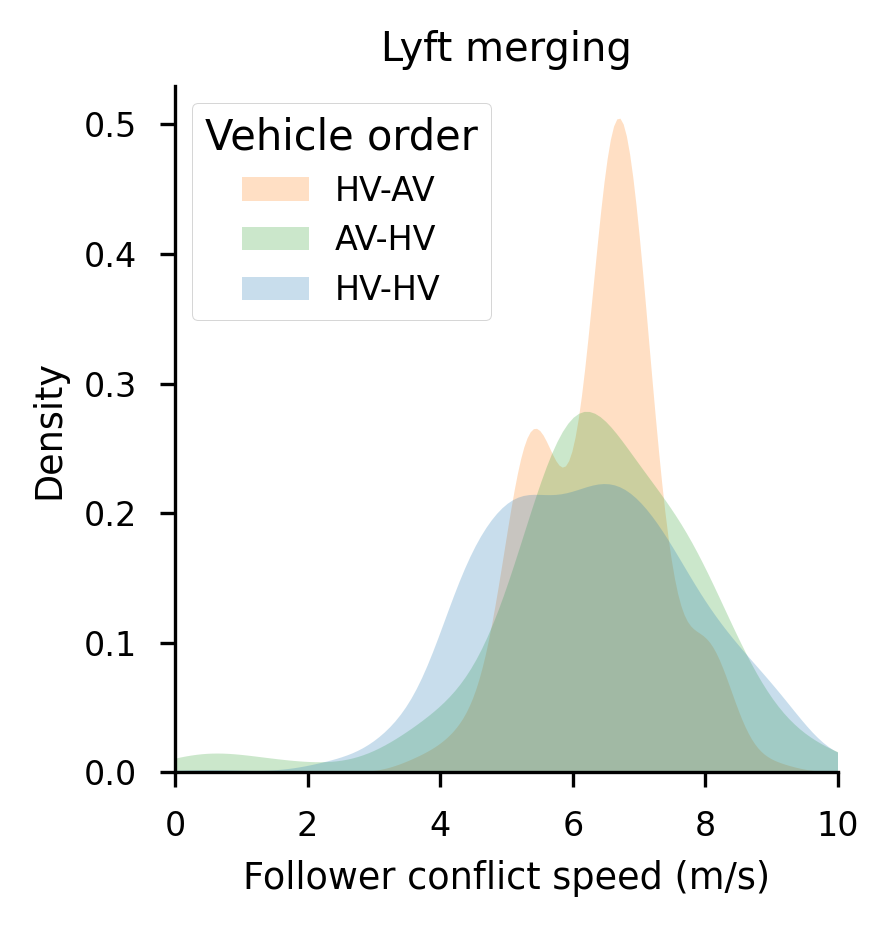}
        \caption{Lyft - Merging}
        \label{fig:sub4}
    \end{subfigure}

    \caption{Distributions of the follower speed at the conflict point}
    \label{fig:conflict_speed}
\end{figure}

The analysis of follower speed at the conflict point, depicted in Figure~\ref{fig:conflict_speed} and Table~\ref{tab:speed_acceleration_comparison} reveals differences not only between AVs and HVs but also between the Waymo and Lyft vehicles. In the Lyft dataset, we observe lower heterogeneity in the speed values, with values similar to or lower than human drivers in similar conditions. Conversely, Waymo's autonomous vehicles exhibited speeds similar to, and even higher than, those of human drivers, with comparable levels of heterogeneity. This finding suggests that Lyft's autonomous vehicles adopt a more conservative approach when navigating unsignalized intersections and entering the conflict zone, while the Waymo vehicle shows a more human-like behavior. These findings highlight the ongoing debate in AV development between prioritizing safety through conservative behavior or maintaining traffic efficiency by mimicking human driving styles. The Lyft approach may lead to potentially increased safety but could negatively impact traffic flow if widely adopted. On the other hand, Waymo's approach may facilitate smoother integration with human-driven traffic but might not fully leverage the potential safety benefits of AVs. 

\begin{table*}
\fontsize{11pt}{12pt}\selectfont
\centering
\caption{\textcolor{black}{Comparison of Speed and Acceleration for Waymo and Lyft datasets in crossing and merging scenarios}}
\renewcommand{\arraystretch}{1.4}  
\resizebox{\textwidth}{!}{
\begin{tabular}{ll cccccc cccccc}
\toprule
& & \multicolumn{6}{c}{Waymo} & \multicolumn{6}{c}{Lyft} \\
\cmidrule(lr){3-8} \cmidrule(lr){9-14}
& & \multicolumn{3}{c}{Crossing} & \multicolumn{3}{c}{Merging} & \multicolumn{3}{c}{Crossing} & \multicolumn{3}{c}{Merging} \\
\cmidrule(lr){3-5} \cmidrule(lr){6-8} \cmidrule(lr){9-11} \cmidrule(lr){12-14}
Test & Benchmark & $\mu$ ($\sigma$) & t-test & u test & $\mu$ ($\sigma$) & t-test & u test & $\mu$ ($\sigma$) & t-test & u test & $\mu$ ($\sigma$) & t-test & u test \\
\midrule
\multicolumn{14}{l}{\textbf{Speed of the Follower (m/s)}} \\
\midrule
\textbf{HV-HV} & & 6.53 (1.36) & & & 7.08 (2.16) & & & 5.94 (1.59) & & & 6.28 (1.65) & & \\
\textbf{AV-HV} & HV-HV & 5.99 (1.26) & \textless0.01 & \textless0.01 & 6.06 (1.82) & \textless0.01 & \textless0.01 & 6.40 (1.41) & 0.03 & \textless0.01 & 6.42 (1.75) & 0.36 & 0.09 \\
\textbf{HV-AV} & HV-HV & 6.28 (1.37) & 0.07 & 0.06 & 6.34 (2.23) & 0.05 & 0.08 & 5.24 (1.52) & \textless0.01 & \textless0.01 & 6.42 (0.99) & 0.26 & 0.08 \\
\textbf{HV-AV} & AV-HV & 6.28 (1.37) & 0.06 & 0.14 & 6.34 (2.23) & 0.40 & 0.39 & 5.24 (1.52) & \textless0.01 & \textless0.01 & 6.42 (0.99) & 0.98 & 0.76 \\
\midrule
\multicolumn{14}{l}{\textbf{Average Speed (m/s)}} \\
\midrule
\textbf{HV-HV} & & 3.40 (0.71) & & & 4.00 (3.76) & & & 3.51 (2.79) & & & 3.66 (1.75) & & \\
\textbf{AV-HV} & HV-HV & 3.20 (0.67) & 0.03 & 0.03 & 3.70 (1.36) & 0.19 & 0.96 & 3.72 (2.75) & 0.37 & 0.40 & 3.90 (2.26) & 0.05 & 0.22 \\
\textbf{HV-AV} & HV-HV & 3.20 (0.40) & 0.01 & 0.02 & 3.50 (1.34) & 0.14 & 0.27 & 2.86 (1.72) & \textless0.01 & \textless0.01 & 3.07 (0.35) & \textless0.01 & \textless0.01 \\
\textbf{HV-AV} & AV-HV & 3.20 (0.40) & 0.79 & 0.82 & 3.50 (1.34) & 0.37 & 0.27 & 2.86 (1.72) & \textless0.01 & \textless0.01 & 3.07 (0.35) & \textless0.01 & \textless0.01 \\
\midrule
\multicolumn{14}{l}{\textbf{Average Acceleration ($\mathbf{\text{m/s}^2}$)}} \\
\midrule
\textbf{HV-HV} & & 1.20 (0.62) & & & 0.60 (0.48) & & & 0.65 (0.66) & & & 0.57 (0.49) & & \\
\textbf{AV-HV} & HV-HV &0.80 (0.62) & \textless0.01 & \textless0.01 & 0.60 (0.66) & 0.61 & 0.50 & 0.83 (0.76) & 0.10 & 0.06 & 0.49 (0.48) & 0.23 & 0.25 \\
\textbf{HV-AV} & HV-HV & 1.30 (0.12) & 0.03 & 0.22 & 1.00 (0.29) & \textless0.01 & \textless0.01 & 1.02 (0.21) & \textless0.01 & \textless0.01 & 0.99 (0.07) & \textless0.01 & \textless0.01 \\
\textbf{HV-AV} & AV-HV & 1.30 (0.12) & \textless0.01 & \textless0.01 & 1.00 (0.29) & \textless0.01 & \textless0.01 & 1.02 (0.21) & 0.07 & 0.61 & 0.99 (0.07) & \textless0.01 & \textless0.01 \\
\bottomrule
\end{tabular}
}
\label{tab:speed_acceleration_comparison}
\end{table*}

\begin{figure}[t]
    \centering
    \begin{subfigure}[b]{0.23\textwidth}
        \centering
        \includegraphics[width=\textwidth]{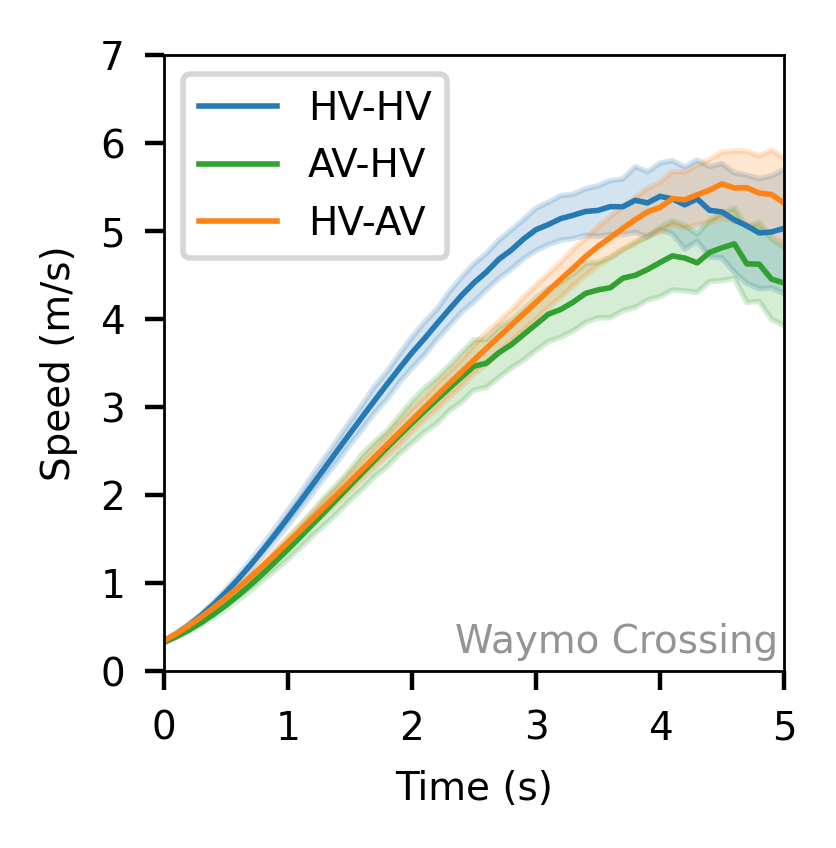}
        \caption{\textcolor{black}{Waymo - Crossing}}
        \label{fig:Waymo - Crossing}
    \end{subfigure}
    \begin{subfigure}[b]{0.23\textwidth}
        \centering
        \includegraphics[width=\textwidth]{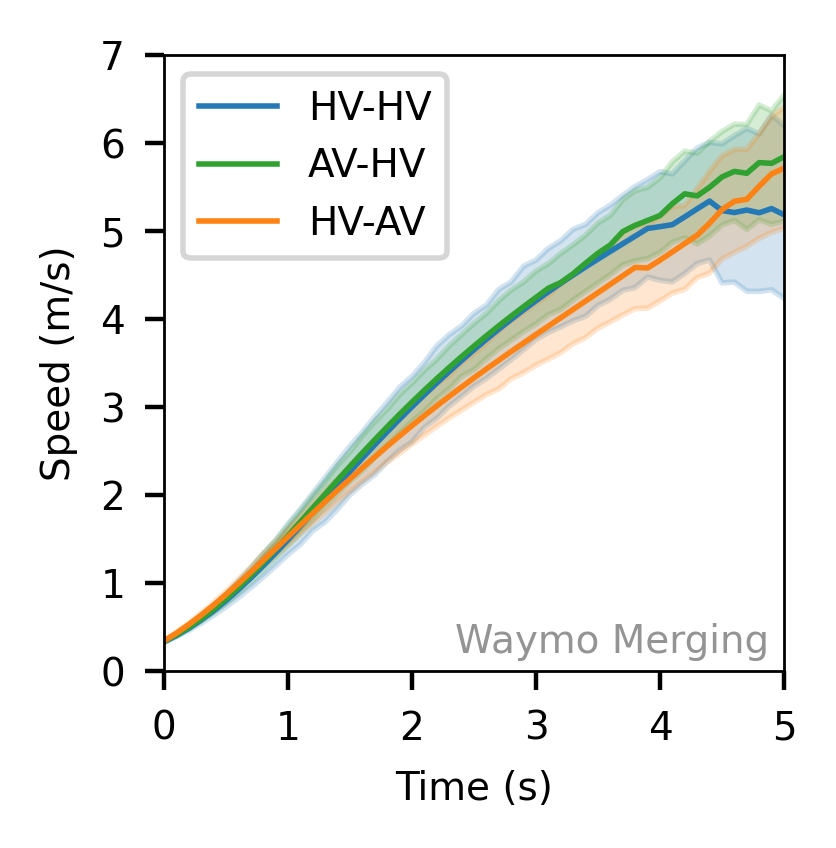}
        \caption{Waymo - Merging}
    \end{subfigure}
    \begin{subfigure}[b]{0.23\textwidth}
        \centering
        \includegraphics[width=\textwidth]{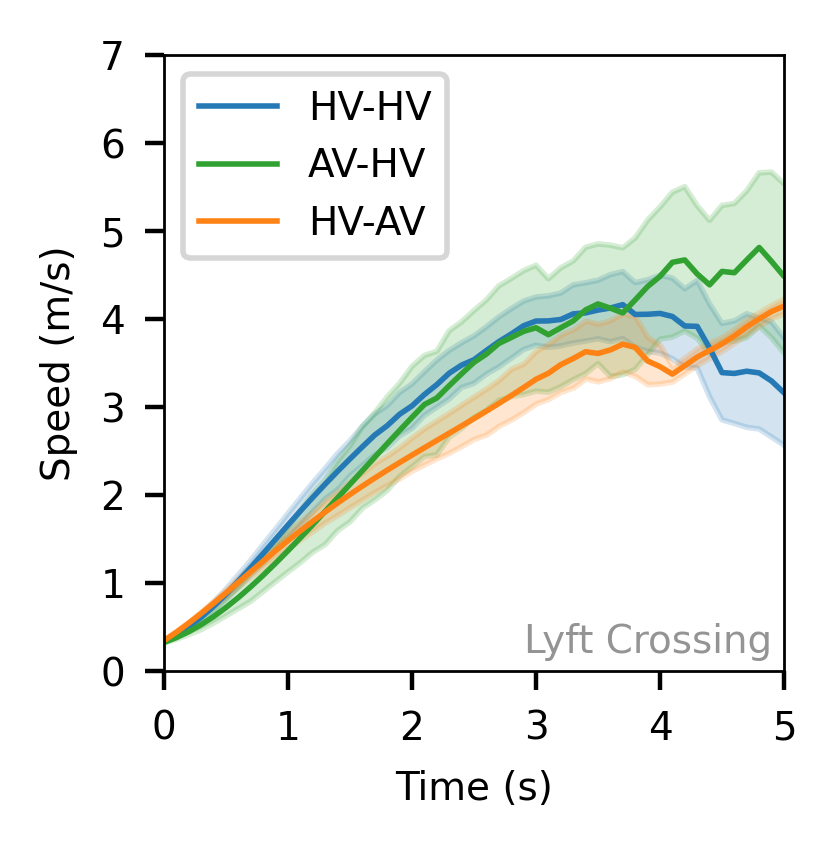}
        \caption{Lyft - Crossing}
    \end{subfigure}
    \begin{subfigure}[b]{0.23\textwidth}
        \centering
        \includegraphics[width=\textwidth]{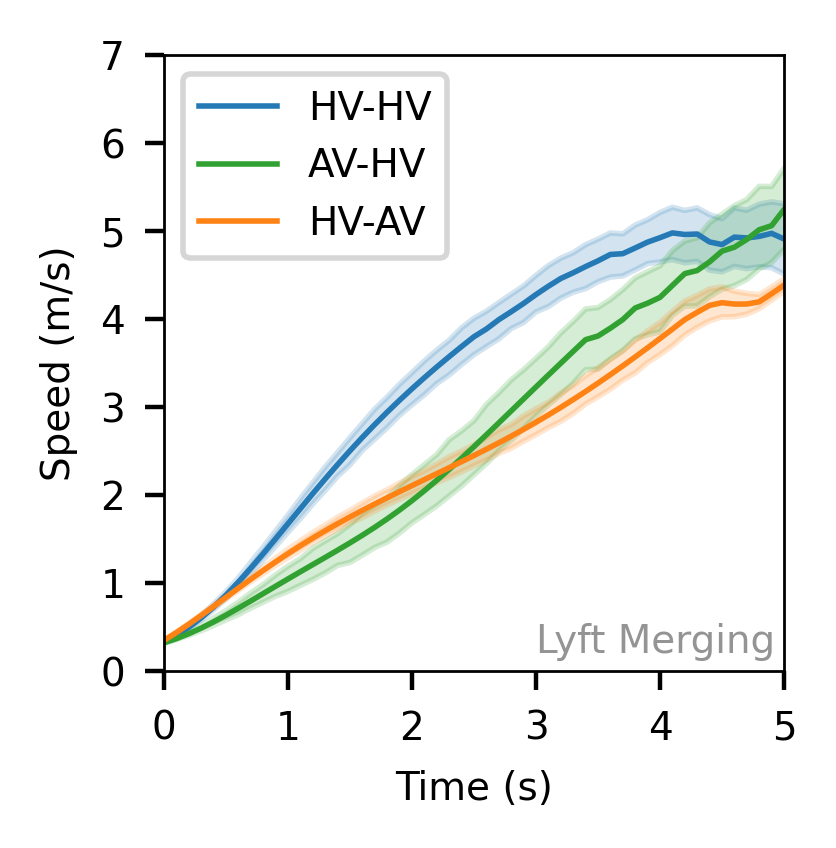}
        \caption{Lyft - Merging}
    \end{subfigure}
    
    \begin{subfigure}[b]{0.23\textwidth}
        \centering
        \includegraphics[width=\textwidth]{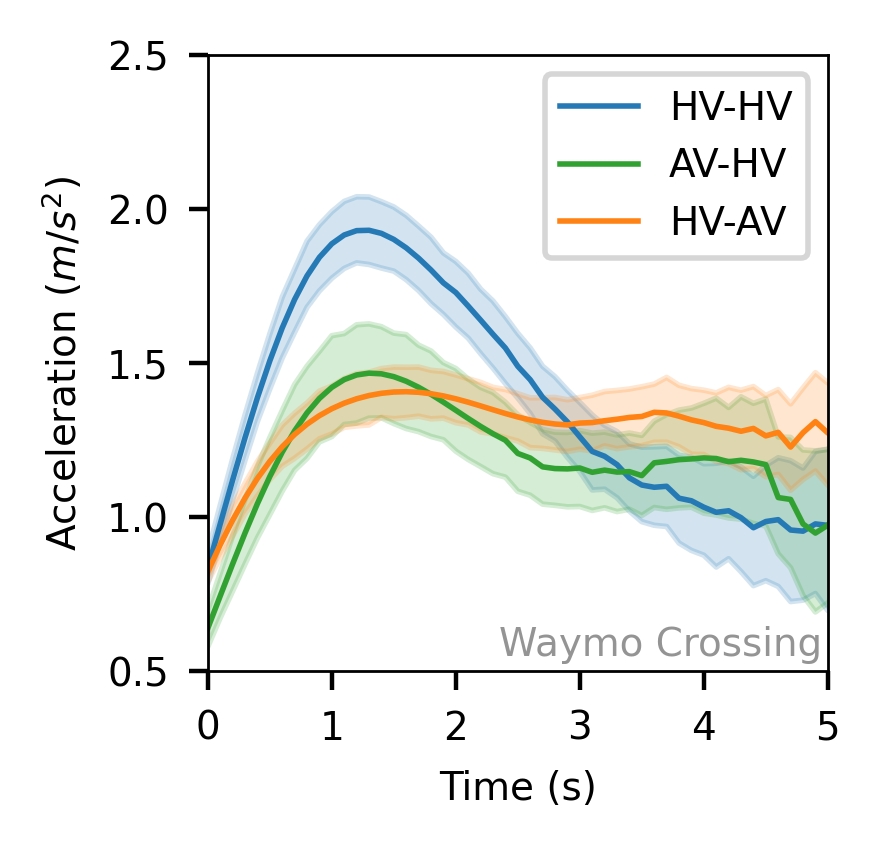}
        \caption{Waymo - Crossing}
        \label{fig:sub5}
    \end{subfigure}
    \begin{subfigure}[b]{0.23\textwidth}
        \centering
        \includegraphics[width=\textwidth]{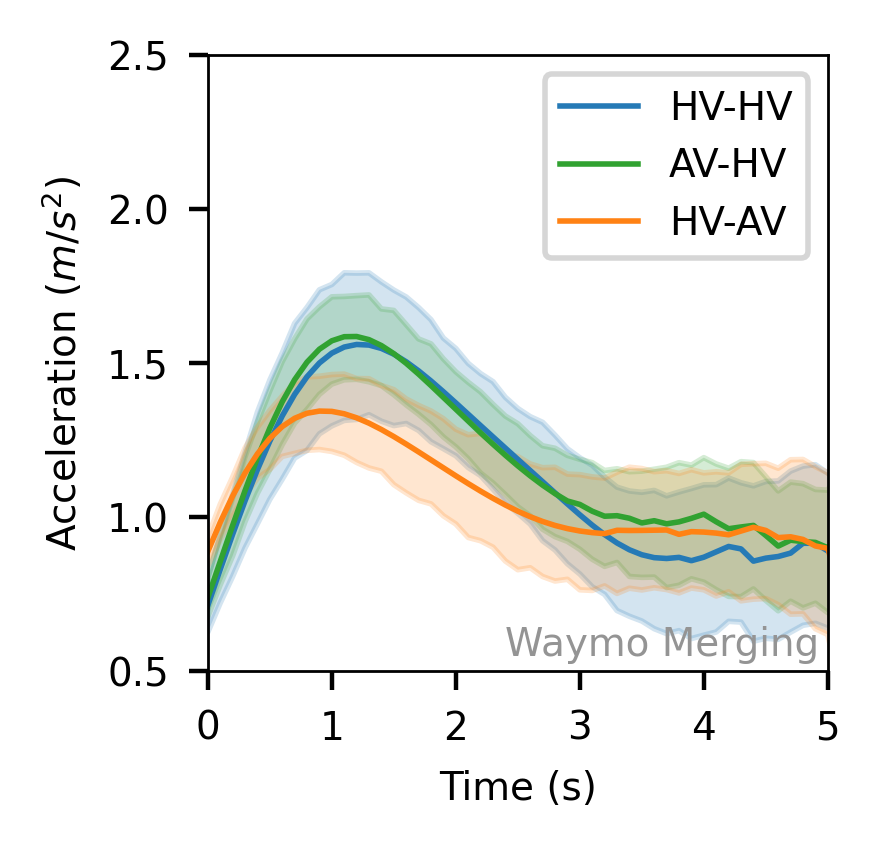}
        \caption{Waymo - Merging}
        \label{fig:sub6}
    \end{subfigure}
    \begin{subfigure}[b]{0.23\textwidth}
        \centering
        \includegraphics[width=\textwidth]{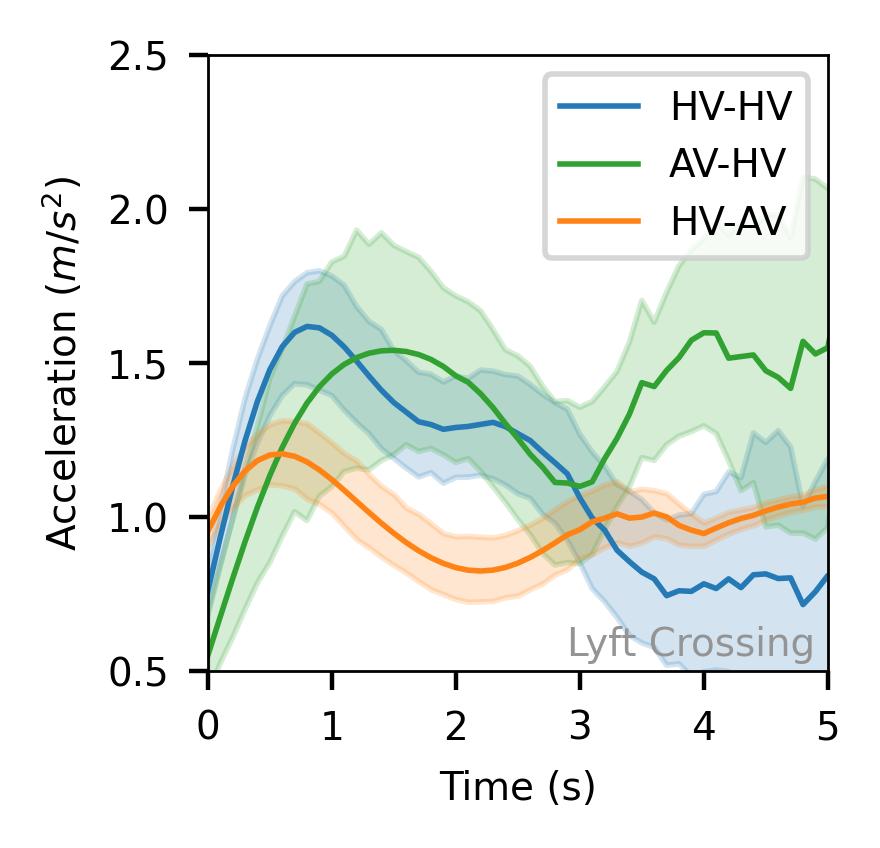}
        \caption{Lyft - Crossing}
        \label{fig:sub7}
    \end{subfigure}
    \begin{subfigure}[b]{0.23\textwidth}
        \centering
        \includegraphics[width=\textwidth]{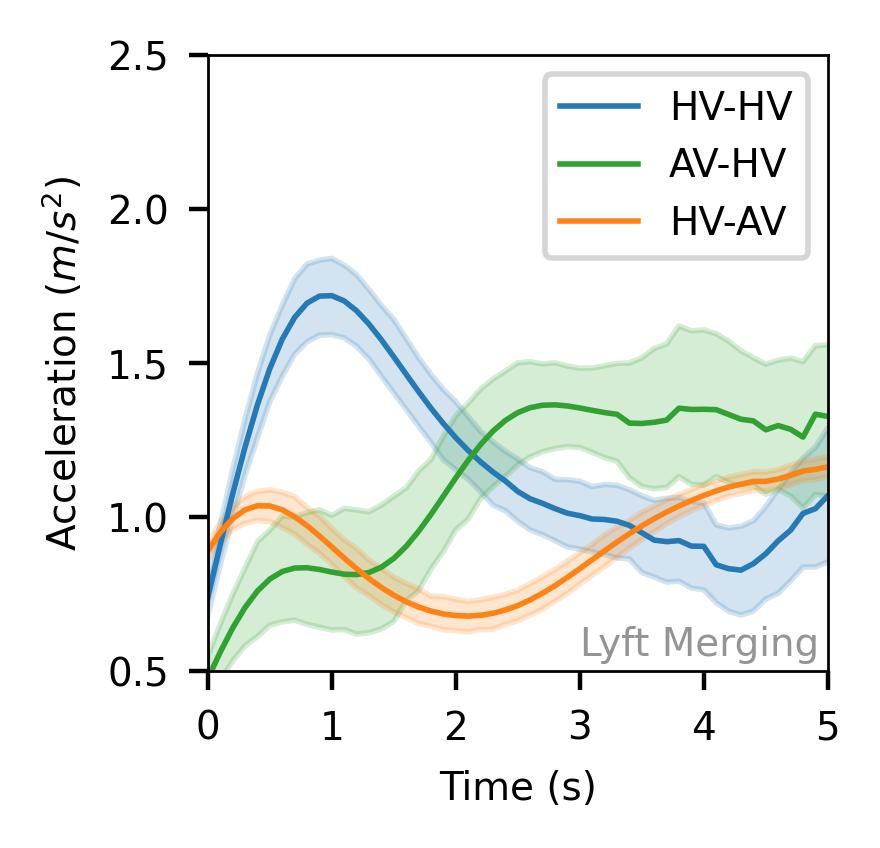}
        \caption{Lyft - Merging}
        \label{fig:sub8}
    \end{subfigure}

    \caption{Speed and Acceleration profiles for Waymo and Lyft datasets}
    \label{fig:speed_profile}
\end{figure}

In the next step, the speed and acceleration profiles of the following vehicle when starting from a standstill are investigated. Figure \ref{fig:speed_profile}  presents these profiles, along with 95\% confidence intervals, and Table \ref{tab:speed_acceleration_comparison} provides the statistical tests evaluating the significance of the observed differences. The results indicate that the speed and acceleration profiles of Waymo vehicles are more closely aligned with those of human drivers, while Lyft vehicles exhibit visibly different patterns. Additionally, the Lyft vehicles show more uniform speed profiles with a gentler slope compared to both human drivers and Waymo vehicles.

Examination of the acceleration profiles yields noteworthy insights as well. Waymo vehicles display an overall acceleration pattern similar to human-driven vehicles, characterized by a rapid initial increase followed by sustained low acceleration. In contrast, Lyft vehicles exhibit a unique acceleration profile. It begins with significantly lower acceleration, experiences a slight decrease midway through the profile, and ends with a mild increase. This unique behavior is probably related to the cautious behavior of the Lyft vehicles to make sure the intersection is safe to pass. Additionally, the acceleration profiles of human-driven vehicles in AV-HV interactions within the Lyft dataset indicate greater variability and uncertainty when interacting with Lyft vehicles. This may stem from the unfamiliar and overly cautious behavior of Lyft AVs, which human drivers might find harder to predict.

Overall, the speed and acceleration analysis from the Waymo and Lyft datasets reveals both contrasting and consistent behaviors. While AVs generally exhibit more uniform and stable behavior, which could improve traffic efficiency, the overly cautious driving style observed in Lyft vehicles raises concerns about whether such conservative behavior might undermine efficiency and provoke unexpected responses from human drivers.

\section{Summary of Findings}
This section summarizes the main findings of this study. The analysis of two large-scale AV datasets revealed that AVs generally maintain larger safety margins compared to HVs, especially when interacting as followers. This is evidenced by higher Post-Encroachment Time and minimum Time to Collision values when AVs trail HVs. This conservative approach by AVs can potentially contribute to increased safety at intersections; however, this benefit might come with the cost of some reduction in traffic efficiency. The findings also suggest that although human drivers may tend to exhibit more consistent and less variable behavior when interacting with AVs, this consistency appears to be influenced by the specific behavior of the AVs themselves. Therefore, a general statement cannot be made across all AV types. The extent to which AVs contribute to standardizing traffic flow patterns and improving safety and efficiency at intersections is likely dependent on how predictable and human-compatible the AV's driving style is.

This study also highlights potential challenges associated with the integration of AVs into mixed-autonomy traffic. Analysis of the maximum required deceleration (MRD) reveals that human drivers following AVs (HV-AV) frequently need to apply more abrupt deceleration rates compared to HV-HV interactions. This observation may be attributed to the unfamiliar or unexpected behavior of AVs, which can potentially lead to misinterpretations and elevated risks for human drivers. These findings highlight a paradox in the analysis of mixed-autonomy traffic flow: while AVs generally maintain larger safety margins, their different driving style can lead to unexpected situations for human drivers, potentially causing unsafe situations. Also, an overly cautious driving style from AVs can lead to aggressive or unsafe driving from human drivers. 

Finally, this study reveals differences between the behavior of Waymo and Lyft vehicles when interacting with HVs at unsignalized intersections. While Lyft AVs demonstrated more conservative behaviors, Waymo AVs exhibited behaviors more similar to human drivers and less conservative compared to the Lyft vehicle. \textcolor{black}{These manufacturer-specific differences have important implications for traffic safety at both individual and system levels. At the individual level, the contrasting approaches create different safety challenges: Lyft's conservative behavior, while theoretically safer, often provokes aggressive responses from human drivers, potentially creating new safety risks. Meanwhile, Waymo's human-like behavior facilitates smoother interactions but may not fully exploit the safety potential of autonomous technology. At the system level, the presence of AVs with different behaviors could create broader safety challenges in mixed-traffic environments. As human drivers encounter AVs from different manufacturers, they may develop inconsistent expectations about AV behavior, leading to confusion and potentially inappropriate responses. This behavioral inconsistency across manufacturers suggests that some degree of standardization in AV behavior could be beneficial for overall traffic safety, particularly during the transition period when both human drivers and different types of AVs share the road. However, determining the optimal standard behavior that balances safety margins with human compatibility remains an open challenge that warrants further research.}

\textcolor{black}{While the datasets used in this study did not include crash data, our findings align with broader observations from crash reports involving automated vehicles collected by the California Department of Transportation \cite{dmv2024}. These reports highlight that unexpected behaviors of automated vehicles are a primary factor in AV-involved accidents \cite{lee2023advancing}, with a significant proportion occurring at or near intersections \cite{boggs2020exploratory, favaro2017examining}. This underscores the critical importance of studying AV and human driver interactions at intersections to improve safety outcomes and minimize potential conflicts in mixed-traffic environments.} Moreover, although the behavior of AVs was shown to be more harmonized and homogeneous in different scenarios compared to HVs, Lyft vehicles showed more consistent and harmonized behaviors compared to Waymo vehicles. These findings underscore the importance of considering manufacturer-specific AV behavior in traffic modeling and management strategies, which is often neglected when evaluating the impacts of AVs on traffic flow efficiency and performance. 

\section{Conclusion}
This study provides an in-depth examination of the interactions between AVs and HVs at unsignalized intersections by utilizing real-world large-scale datasets from Waymo and Lyft. The research underscores the intricate dynamics present in mixed-autonomy traffic environments, highlighting the importance of understanding the mutual influence of AVs and human drivers. By evaluating key safety metrics such as Time to Collision (TTC), Post-Encroachment Time (PET), Maximum Required Deceleration (MRD), and Time Advantage (TA), this study sheds light on the behavioral differences and adaptations between AVs and HVs. One of the key findings of this study is that AVs tend to maintain larger safety margins than HVs, which enhances safety but may also reduce traffic efficiency. However, the research reveals that human drivers exhibit more consistent behavior when interacting with AVs, suggesting that AVs could standardize traffic flow patterns at intersections. This potential for harmonization indicates a positive influence of AVs on overall traffic dynamics. The study further identifies significant differences between the behaviors of Waymo and Lyft AVs. Waymo vehicles tend to mimic human driving behaviors more closely, leading to smoother integration with human traffic. In contrast, Lyft vehicles display more conservative driving patterns, which may increase safety but at the cost of potential inefficiencies in traffic flow. These differences underscore the importance of considering manufacturer-specific differences in traffic modeling and management strategies. The study also highlights potential challenges in mixed-autonomy traffic. The unexpected behaviors of AVs can lead to misunderstandings and increased risk for human drivers, as indicated by higher MRD values in AV-HV interactions. This paradox of maintaining safety margins while potentially causing unsafe situations due to unpredictability underscores the need for improved AV algorithms that align more closely with human expectations and reasoning.

While this study provides valuable insights into the interactions between AVs and HVs at unsignalized intersections, several limitations must be acknowledged. One limitation of this study is the use of datasets collected in different cities, which may affect the driving behaviors of both human drivers and AVs. Although we took measures to minimize this impact by using the data in the same country and state, doing mostly relative comparisons, and using rigorous statistical tests, future work should consider the utilization of datasets collected in similar geographical locations upon availability. Moreover, although we’ve implemented measures such as comprehensive data preprocessing, multi-metric analysis, use of diverse datasets, consideration of behavioral context, and rigorous statistical testing to address potential biases highlighted in \cite{jiao2024beyond}, some limitations may persist. These include residual biases from sensor noise and limited representation of AV behavior diversity, which should be considered when interpreting and applying the results of this study. Last but not least, in this study, we focused on non-signalized intersections with equal priority at all approaches. The analysis of AV-HV interactions at prioritized intersections is another interesting line of research, which can be followed in future works.




\begin{ac}
The authors confirm their contribution to the paper as follows: study conception and design; analysis and interpretation of results: S. Rahmani, Z. Xu, S.C. Calvert, and B. van Arem; draft manuscript preparation: S. Rahmani, Z. Xu, and S.C. Calvert. All authors reviewed the results and approved the final version of the manuscript.
\end{ac}

\begin{dci}
The authors declared no potential conflicts of interest with respect to the research, authorship, and/or publication of this article.
\end{dci}

\begin{funding}
This project has received funding from the European Union's Horizon 2020 research and innovation programme under grant agreement No 101006664. The sole responsibility of this publication lies with the authors. Neither the European Commission nor CINEA –- in its capacity of Granting Authority –- can be made responsible for any use that may be made of the information this document contains. The authors would like to thank all partners within the Hi-Drive project for their cooperation and valuable contribution.
\end{funding}

\begin{das}
The processed dataset used in this study, alongside with the developed algorithms and scripts are openly accessible at \url{https://github.com/SaeedRahmani/Unsignalized_AV_HV}
\end{das}

\bibliographystyle{TRR}
\bibliography{main.bib}

\end{document}